\documentclass[sigconf]{acmart}
\AtBeginDocument{%
  }

\copyrightyear{2022} 
\acmYear{2022} 
\setcopyright{acmcopyright}
\acmConference[KDD '22] {Proceedings of the 28th ACM SIGKDD Conference on Knowledge Discovery and Data Mining}{August 14--18, 2022}{Washington, DC, USA.}
\acmBooktitle{Proceedings of the 28th ACM SIGKDD Conference on Knowledge Discovery and Data Mining (KDD '22), August 14--18, 2022, Washington, DC, USA}
\acmPrice{15.00}
\acmISBN{978-1-4503-9385-0/22/08}
\acmDOI{10.1145/3534678.3539337}
\usepackage{algorithm}
\usepackage{algpseudocode}
\usepackage{amsmath}
\usepackage{mathtools}
\usepackage{subcaption}
\usepackage{wrapfig}
\usepackage{xcolor}
\usepackage{booktabs}
\usepackage{xspace}
\usepackage{multirow}
\usepackage{stmaryrd}
\usepackage{mdwlist}
\usepackage{enumitem}

\newcommand{\method}[0]{SVGA\xspace}
\newcommand{\methodlong}[0]{Structured Variational Graph Autoencoder\xspace}

\newcommand{\xall}[0]{\mathbf{X}}
\newcommand{\yall}[0]{\mathbf{y}}
\newcommand{\xset}[0]{\mathcal{V}_{x}}
\newcommand{\yset}[0]{\mathcal{V}_{y}}
\newcommand{\zvar}[0]{\mathbf{Z}}
\newcommand{\adj}[0]{\mathbf{A}}
\newcommand{\elbo}[0]{\mathcal{L}(\Theta)}
\newcommand{\zdist}[0]{q_\phi (\zvar \mid \xall, \yall, \adj)}
\newcommand{\prior}[0]{p (\zvar \mid \adj)}

\algrenewcommand\algorithmicrequire{\textbf{Input:}}
\algrenewcommand\algorithmicensure{\textbf{Output:}}

\DeclarePairedDelimiterX{\infdivx}[2]{(}{)}{%
  #1\;\delimsize|\delimsize|\;#2%
}
\newcommand{\kld}[2]{\ensuremath{D_\mathrm{KL}\infdivx{#1}{#2}}}

\numberwithin{problem}{section}

\usepackage{caption}
\usepackage{array}
\newcommand{\PreserveBackslash}[1]{\let\temp=\\#1\let\\=\temp}
\newcolumntype{C}[1]{>{\PreserveBackslash\centering}p{#1}}
\newcolumntype{R}[1]{>{\PreserveBackslash\raggedleft}p{#1}}
\newcolumntype{L}[1]{>{\PreserveBackslash\raggedright}p{#1}}

\begin{document}

\title{Accurate Node Feature Estimation with Structured Variational Graph Autoencoder}

\author{Jaemin Yoo}
\authornote{This work was done when the author was at Seoul National University.}
\affiliation{%
  \institution{Carnegie Mellon University}
  \city{Pittsburgh}
  \state{PA}
  \country{USA}
}
\email{jaeminyoo@cmu.edu}

\author{Hyunsik Jeon}
\affiliation{%
  \institution{Seoul National University}
  \city{Seoul}
  \country{South Korea}
}
\email{jeon185@snu.ac.kr}

\author{Jinhong Jung}
\affiliation{%
  \institution{JBNU}
  \city{Jeonju}
  \country{South Korea}
}
\email{jinhongjung@jbnu.ac.kr}

\author{U Kang}
\affiliation{%
  \institution{Seoul National University}
  \city{Seoul}
  \country{South Korea}
}
\email{ukang@snu.ac.kr}

\renewcommand{\shortauthors}{Jaemin Yoo et al.}

\begin{abstract}
Given a graph with partial observations of node features, how can we estimate the missing features accurately?
Feature estimation is a crucial problem for analyzing real-world graphs whose features are commonly missing during the data collection process.
Accurate estimation not only provides diverse information of nodes but also supports the inference of graph neural networks that require the full observation of node features.
However, designing an effective approach for estimating high-dimensional features is challenging, since it requires an estimator to have large representation power, increasing the risk of overfitting.
In this work, we propose \method (\methodlong), an accurate method for feature estimation.
\method applies strong regularization to the distribution of latent variables by structured variational inference, which models the prior of variables as Gaussian Markov random field based on the graph structure.
As a result, \method combines the advantages of probabilistic inference and graph neural networks, achieving state-of-the-art performance in real datasets.

\end{abstract}



%\keywords{node feature estimation, graph neural networks, Gaussian Markov random fields, structured variational inference, graph Laplacian}


\maketitle

\section{Introduction}

\emph{Given a graph with partial observations of node features, how can we estimate the missing features accurately?}
Many real-world data are represented as graphs to model the relationships between entities.
Social networks, seller-item graphs in electronic commerce, and user-movie graphs in a streaming service are all examples of graph data that have been studied widely in literature \cite{Yoo2017, Kipf2017, Velickovic2018, Shchur2018, Hu2020}.
Such graphs become more powerful when combined with feature vectors that describe the diverse properties of nodes \cite{Duong2019, Yang2020}.

\begin{figure}
	\centering
	\vspace{2mm}
	\includegraphics[width=0.47\textwidth]{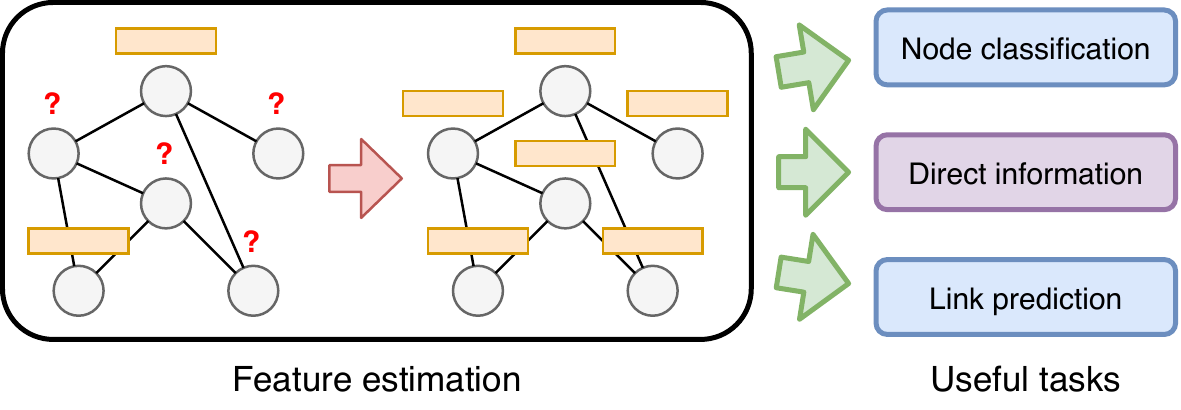}
	\caption{
		An illustration of the feature estimation problem.
		The generated features not only provide direct information of node properties but also help other graph-related tasks.
	}
\label{fig:motivation}
\end{figure}

However, node features are commonly missing in a real-world graph.
Users in an online social network set their profiles private, and sellers in electronic commerce often register items without an informative description.
In such cases, even the observed features cannot be used properly due to the missing ones, since many graph algorithms assume the full observation of node features.
Figure~\ref{fig:motivation} illustrates the feature estimation problem in an example graph.
An accurate estimation of missing features not only provides diverse information of node properties but also improves the performance of essential tasks such as node classification or link prediction by providing important evidence for training a classifier.

However, accurate estimation of missing features is challenging due to the following reasons.
First, target nodes have no specific information that describes their properties.
The main evidence for estimation is the graph structure, which gives only partial information of nodes based on the relationships with the other nodes.
Second, the target variables are high-dimensional vectors containing up to thousands of elements.
This requires large representation power for accurate estimation, involving a high risk of overfitting as a consequence.
Existing approaches \cite{Huang2019, Chen2019, Chen2020} failed to address such challenges effectively, resulting in limited performance.

We propose \method (\methodlong), an accurate method for missing feature estimation.
The main idea for addressing the challenges is to run \emph{structured} variational inference to effectively regularize the distribution of latent variables by modeling their correlations from the structure of a graph.
We first propose stochastic inference, which models the prior of latent variables as Gaussian Markov random field (GMRF).
%The main idea for addressing the challenges of problem is to run \emph{structured} variational inference without ignoring the correlations between target variables.
%We first propose stochastic inference with modeling the prior of latent variables as Gaussian Markov random field (GMRF), using the graph structure as an effective regularizer for the distribution of node representations.
Then, we improve the stability of inference with our proposed deterministic modeling, which results in a new graph-based regularizer.
These allow us to avoid the overfitting without degrading the representation power, achieving state-of-the-art performance in real-world datasets.

Our contributions are summarized as follows:
\begin{itemize}
	\item \textbf{Method.}
		We propose \method, an accurate method for missing feature estimation.
		\method introduces a new way to run variational inference on graph-structured data with modeling the correlations between target variables as GMRF.
	\item \textbf{Theory.}
		We analyze the theoretical properties of structured variational inference with the stochastic and deterministic modeling.
		We also analyze the time and space complexities of our \method, which are both linear with the number of nodes and edges of a given graph, showing its scalability.
	\item \textbf{Experiments.}
		Extensive experiments on eight real-world datasets show that \method provides state-of-the-art performance with up to 16.3\% higher recall and 14.0\% higher nDCG scores in feature estimation, and up to 14.2\% higher accuracy in node classification compared to the best competitors.
\end{itemize}

The rest of this paper is organized as follows.
In Section \ref{sec:preliminaries}, we introduce the problem definition and preliminaries of \method.
In Section \ref{sec:proposed-approach}, we propose \method and discuss its theoretical properties.
We present experimental results in Section \ref{sec:experiments} and describe related works in Section \ref{sec:related-works}.
We conclude in Section \ref{sec:conclusion}.
The code and datasets are available at \underline{\smash{\url{https://github.com/snudatalab/SVGA}}}.

\section{Preliminaries}
\label{sec:preliminaries}

We introduce the problem definition and preliminaries, including Gaussian Markov random field and variational inference.
%Symbols used frequently in this paper are summarized in Appendix \ref{appendix:symbols}.

\subsection{Missing Feature Estimation}
\label{subsec:problem-definition}

The feature estimation problem is defined as follows.
We have an undirected graph $G = (\mathcal{V}, \mathcal{E})$, where $\mathcal{V}$ and $\mathcal{E}$ represent the sets of nodes and edges, respectively.
A feature vector $\mathbf{x}_i$ exists for every node $i$, but is observable only for a subset $\mathcal{V}_x \subset \mathcal{V}$ of nodes.
Our goal is to predict the missing features of \emph{test} nodes $\mathcal{V} \setminus \mathcal{V}_x$ using the structure of $G$ and the observations for $\mathcal{V}_x$.
The problem differs from generative learning \cite{Kingma2014} in that there exist correct answers; generative learning is typically an unsupervised problem.

We also assume that the label $y_i$ of each node $i$ can be given as an additional input for a set $\mathcal{V}_y$ of nodes such that $\mathcal{V}_y \subseteq \mathcal{V}$.
Such labels improve the accuracy of feature estimation, especially when they provide information for the test nodes: $\mathcal{V}_y \cap (\mathcal{V} \setminus \mathcal{V}_x) \neq \emptyset$.
This is based on the idea that categorical labels are often easier to acquire than high-dimensional features, and knowing the labels of target nodes gives a meaningful advantage for estimation.
Thus, we design our framework to be able to work with $\mathcal{V}_y \neq \emptyset$, although we consider $\mathcal{V}_y = \emptyset$ as a base setup of experiments for the consistency with previous approaches that take only the observed features.

%\textbf{Evaluation.}
%We evaluate the performance of feature estimation in two ways.
%First, we directly compare predictions with the true features that are unknown at the training time.
%Second, we solve the node classification problem utilizing the generated features to quantify how well they model the relationships between nodes in terms of labels.
%This is based on the idea that generated features can be informative for solving other tasks regardless of the error from the true features.
%Detailed information is in Section \ref{ssec:exp-setup}.

\subsection{Gaussian Markov Random Field}

\begin{figure}
	\centering
	\includegraphics[width=0.425\textwidth]{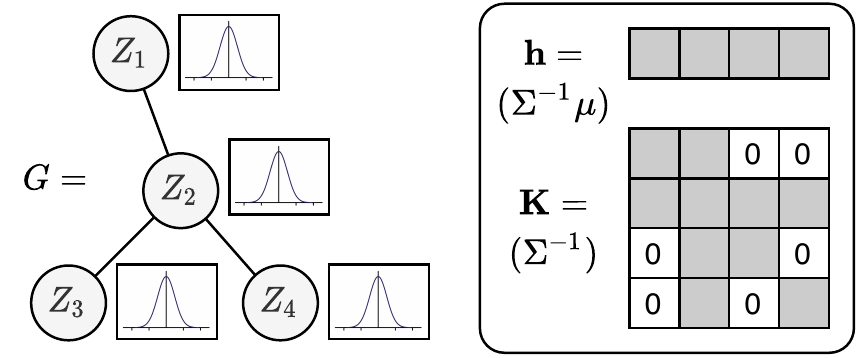}
	\caption{
		Gaussian Markov random field (GMRF) describing a Gaussian distribution $\mathcal{N}(\mu, \Sigma)$ by parameters $\mathbf{h}$ and $\mathbf{K}$.
		The nonzero entries in $\mathbf{K}$ correspond to the edges in $G$.
	}
	\label{fig:gmrf}
\end{figure}

\begin{figure*}
	\centering
	\includegraphics[width=0.9\textwidth]{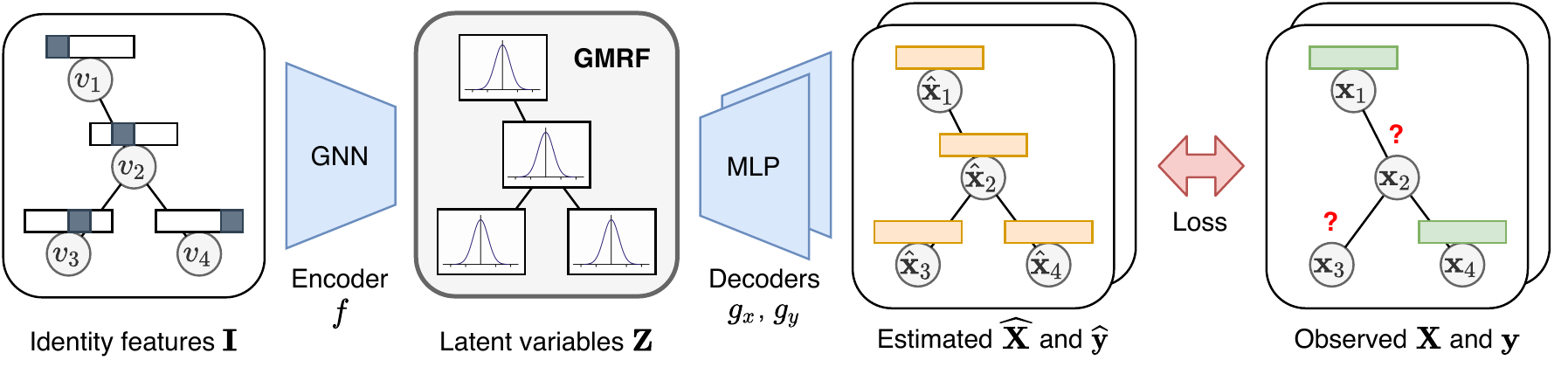}
	\caption{%
		The structure of our \method, which consists of an encoder network $f$ and two decoder networks $g_x$ and $g_y$ for features and labels, respectively.
		We model the distribution of latent variables with GMRF, exploiting the graph structure for modeling the correlations between target variables.
		The label decoder $g_y$ works as an auxiliary module that helps $g_x$.
	}
	\vspace{-3mm}
	\label{fig:overview}
\end{figure*}

Gaussian Markov random field (GMRF) \cite{Koller2009} is a graphical model that represents a multivariate Gaussian distribution.
Given a graph $G = (\mathcal{V}, \mathcal{E})$ whose nodes have continuous signals that are correlated by the graph structure, GMRF represents the distribution of signals with two kinds of potential functions $\psi_i$ and $\psi_{ij}$ for every node $i$ and edge $(i, j)$, respectively.
We assume the signal of each node $i$ as a random variable $Z_i$ with a possible value $z_i$.

Specifically, the node potential $\psi_i$ for each node $i$ and the edge potential $\psi_{ij}$ for each edge $(i, j)$ are defined as follows:
\begin{align}
	& \psi_i(z_i) = \exp(-0.5K_{ii}z_i^2 + h_i z_i) \label{eq:mrf-np} \\
	& \psi_{ij}(z_i, z_j) = \exp(-K_{ij}z_iz_j), \label{eq:mrf-ep}
\end{align}
where $\mathbf{h} \in \mathbb{R}^n$ and $\mathbf{K} \in \mathbb{R}^{n \times n}$ are the parameters of the GMRF, and $n$ is the number of nodes.
The nonzero elements of $\mathbf{K}$ correspond to the edges of the graph as depicted in Figure \ref{fig:gmrf}.

Then, the joint probability $p(\mathbf{z})$ is given as the multiplication of all potential functions:
\begin{equation}
	p(\mathbf{z}) = \frac{1}{C} \prod_{i \in \mathcal{V}} \psi_i(z_i) \prod_{(i, j) \in \mathcal{E}} \psi_{ij}(z_i, z_j),
	\label{eq:mrf-joint}
\end{equation}
where $C$ is a normalization constant.
Each potential measures how likely $z_i$ or $(z_i, z_j)$ appears with the current probabilistic assumption with the parameters $\mathbf{h}$ and $\mathbf{K}$, and the joint probability is computed by multiplying the potentials for all nodes and edges.

The roles of parameters $\mathbf{K}$ and $\mathbf{h}$ can be understood with respect to the distribution that GMRF represents.
Lemma \ref{lemma:gmrf} shows that GMRF is equivalent to a multivariate Gaussian distribution whose mean and covariance are determined by $\mathbf{K}$ and $\mathbf{h}$.
$\mathbf{K}$ is the inverse of the covariance $\Sigma$, and a pair of signals $z_i$ and $z_j$ is more likely to be observed if $K_{ij}$ is small.
$\mathbf{h}$ determines the mean of the signals if $\mathbf{K}$ is fixed, and is typically set to zero as we assume no initial bias of signals for the simplicity of computation.

\begin{lemma}
	The joint probability of Equation \eqref{eq:mrf-joint} is the same as the probability density function of a multivariate Gaussian distribution $\mathcal{N}(\mu, \Sigma)$, where $\mu = \mathbf{K}^{-1}\mathbf{h}$ and $\Sigma = \mathbf{K}^{-1}$.
	\label{lemma:gmrf}
\end{lemma}

\begin{proof}
	\vspace{-1mm}
	See Appendix \ref{appendix:proof-gmrf}.
	\vspace{-1mm}
\end{proof}

We utilize GMRF to incorporate a real-world graph in a probabilistic framework.
Specifically, we generate a multivariate Gaussian distribution that models the probabilistic relationships between nodes by designing $\mathbf{K}$ and $\mathbf{h}$ from the adjacency matrix $\mathbf{A}$ of the given graph.
GMRF plays a crucial role in our proposed approach, which aims to run variational inference in graph-structured data without ignoring the correlations between target variables.

\subsection{Variational Inference for Joint Learning}

Variational inference \cite{Kingma2014, Kipf2016, Tomczak2020} is a technique for approximating intractable posterior distributions, which has been used widely for generative learning.
Given the adjacency matrix $\adj$ of a graph, our goal is to find optimal parameters $\Theta$ that maximize the likelihood $p_\Theta(\xall, \yall \mid \adj)$ of observed features $\xall$ and labels $\yall$.
We introduce a latent variable $\mathbf{z}_i \in \mathbb{R}^d$ for each node $i$ and denote the realization of all latent variables by $\mathbf{Z} \in \mathbb{R}^{n \times d}$, where $n$ is the number of nodes and $d$ is the size of variables.
The latent variable $\mathbf{z}_i$ represents the characteristic of each node $i$ for estimating its feature $\mathbf{x}_i$.

With variational inference, we change the problem into maximizing the evidence lower bound (ELBO):
\begin{equation}
\begin{split}
	&\log p_\Theta(\xall, \yall \mid \adj) \geq \elbo \\
	&\quad \quad = \mathbb{E}_{\zvar \sim \zdist} [\log p_{\theta, \rho} (\xall, \yall \mid \zvar, \adj)] \\
	&\quad \quad \quad \quad \quad \quad \quad \quad - \kld{\zdist}{p(\zvar \mid \adj)},
\end{split}
\label{eq:elbo-1}
\end{equation}
where $\elbo$ is the ELBO, $q_\phi$ is a parameterized distribution of $\zvar$, and $p_{\theta, \rho}$ is a parameterized distribution of $\xall$ and $\yall$.
The first term of $\elbo$ is the likelihood of observed variables given $\zvar$, while the second term measures the difference between $\zdist$ and the prior distribution $p(\zvar \mid \adj)$ by the KL divergence.

We assume the conditional independence between $\xall$, $\yall$, and $\adj$ given $\zvar$, expecting that each variable $\mathbf{z}_i$ has sufficient information of node $i$ to generate its feature $\mathbf{x}_i$ and label $y_i$.
Then, the first term of $\elbo$ in Equation \eqref{eq:elbo-1} is rewritten as follows:
\begin{multline}
	\mathbb{E}_{\zvar \sim q_\phi (\zvar \mid \xall, \yall, \adj)} [\log p_{\theta, \rho} (\xall, \yall \mid \zvar, \adj)] \\
	= \mathbb{E}_{\zvar \sim q_\phi (\zvar \mid \cdot)} \Bigl[\sum_{i \in \xset} \log p_\theta (\mathbf{x}_i \mid \mathbf{z}_i) + \sum_{i \in \yset} \log p_\rho (y_i \mid \mathbf{z}_i) \Bigr],
\label{eq:elbo-3}
\end{multline}
where $\xset$ and $\yset$ are the sets of nodes whose features and labels are observed, respectively, and $q_\phi (\zvar \mid \cdot)$ denotes $q_\phi (\zvar \mid \xall, \yall, \adj)$.

Equation \eqref{eq:elbo-3} represents the conditional likelihood of observed features and labels given $\mathbf{Z}$.
Thus, maximizing Equation \eqref{eq:elbo-3} is the same as minimizing the reconstruction error of observed variables in typical autoencoders.
On the other hand, the KL divergence term in Equation \eqref{eq:elbo-1} works as a regularizer that forces the distribution $\zdist$ of latent variables to be close to the prior $p(\zvar \mid \adj)$.
The characteristic of regularization depends on how we model the prior $p(\zvar \mid \adj)$, which plays an essential role in our framework.

Note that the objective function of Equation \eqref{eq:elbo-1} works whether the observed labels $\yall$ are given or not, due to our assumption on the conditional independence between $\xall$ and $\yall$.
Only the first term of Equation \eqref{eq:elbo-3} is used if there are no observed labels.

\section{Proposed Method}
\label{sec:proposed-approach}

We propose \method (\methodlong), an accurate method for missing feature estimation.
The main ideas of \method are summarized as follows:
\begin{itemize}
	\item \textbf{GNN with identity node features (Sec. \ref{ssec:method-gnn-modeling}).}
		We address the deficiency of input features by utilizing a graph neural network (GNN) with identity node features as an encoder function, which allows us to learn an independent embedding vector for each node during the training.
	\item \textbf{Structured variational inference (Sec. \ref{ssec:method-variational-inference}).}
		We propose a new way to run variational inference on graph-structured data without ignoring the correlations between target examples.
		This is done by modeling the prior distribution of latent variables with Gaussian Markov random field (GMRF).
	\item \textbf{Unified deterministic modeling (Sec. \ref{ssec:deterministic-modeling}).}
		We improve the stability of inference by changing the stochastic sampling of latent variables into a deterministic process.
		This makes the KL divergence term of ELBO as a general regularizer that controls the space of node representations.
\end{itemize}

In Section \ref{ssec:method-gnn-modeling}, we introduce the overall structure of \method and the objective function for its training.
Then in Sections \ref{ssec:method-variational-inference} and \ref{ssec:deterministic-modeling}, we induce our graph-based regularizer from structured variational inference.
Specifically, we propose the basic parameterization of structured inference in Section \ref{ssec:method-variational-inference} and improve its stability with the deterministic modeling of latent variables in Section \ref{ssec:deterministic-modeling}.

\subsection{Overall Structure of \method}
\label{ssec:method-gnn-modeling}

Figure \ref{fig:overview} shows the overall structure of \method, which consists of an encoder $f$ and two decoder networks $g_x$ and $g_y$.
The networks $f$, $g_x$ and $g_y$ are designed to estimate the target distributions of the ELBO of Equation \eqref{eq:elbo-1}: $\zdist$, $p_\theta(\mathbf{x}_i \mid \mathbf{z}_i)$ and $p_\rho (y_i \mid \mathbf{z}_i)$, respectively, where $\phi$, $\theta$, and $\rho$ are their parameters.
The encoder $f$ generates latent representations of nodes, and the decoders $g_x$ and $g_y$ use the generated representations to estimate the features and labels of nodes.
The feature decoder $g_x$ makes the final estimation of missing features, while the label decoder $g_y$ helps the training of $g_x$ and is not used if no labels are observed.

\subsubsection{Encoder Network}

The encoder network $f$ aims to model the latent distribution $\zdist$ with parameters $\phi$.
We propose to use a graph neural network (GNN) as $f$, because the main functionality required for $f$ is to generate an embedding vector for each node following the graphical structure.
In experiments, we adopt a simple graph convolutional network (GCN) \cite{Kipf2017} as $f$, which works well even when the amount of training data is insufficient.

Still, it is required that every node contains a feature vector to run the GNN encoder on the given graph.
Only a few nodes have observed features in our case, and it makes an imbalance between nodes with and without observed features.
Thus, we use the identity matrix $\mathbf{I} \in \mathbb{R}^{n \times n}$ as the input of $f$, using the observed features only as the answer for the training of \method.
This allows $f$ to learn an independent embedding for each node at its first layer and to have sufficient capacity to generate diverse node representations.
%\footnote{\blue{We tried various types of initial features such as random normal vectors or structural features made from the adjacency matrix, but the identity matrix worked the best.}}

If we use a GCN with two layers as in previous work \cite{Kipf2017}, the encoder function $f$ is defined as $f(\adj; \phi) = \hat{\adj} (\sigma(\hat{\adj} \mathbf{I} \mathbf{W}_1)) \mathbf{W}_2$, where {\small $\hat{\mathbf{A}} = \tilde{\mathbf{D}}^{-1/2} \tilde{\mathbf{A}} \tilde{\mathbf{D}}^{-1/2}$} is the normalized adjacency matrix, $\tilde{\mathbf{A}} = \mathbf{A} + \mathbf{I}$ is the adjacency matrix with self-loops, $\tilde{\mathbf{D}}$ is the degree matrix such that {\small $\tilde{D}_{ii} = \sum_j \tilde{A}_{ij}$, and $\sigma$} is the ReLU function.
{\small $\mathbf{W}_1 \in \mathbb{R}^{n \times d}$} and {\small $\mathbf{W}_2 \in \mathbb{R}^{d \times d}$} are the weight matrices of layers 1 and 2, respectively, where $n$ is the number of nodes, and $d$ is the size of latent variables.
We do not represent the bias terms for brevity.
Note that the node feature matrix of the original formulation of GCN \cite{Kipf2017} is replaced with the identity matrix $\mathbf{I}$ based on our idea of identity node features.

\begin{algorithm}[t]
\caption{
	Training of \method with deterministic inference.
}
\begin{algorithmic}[1]
	\Require Adjacency matrix $\mathbf{A}$, diagonal adjacency $\mathbf{D}$, feature $\mathbf{X}$, (optional) one-hot label $\mathbf{Y}$, hyperparameters $\alpha$, $\beta$ and $\lambda$, networks $f$, $g_x$, and $g_y$, and their parameters $\phi$, $\theta$, and $\rho$, respectively
	\Ensure Updated parameters $\phi'$, $\theta'$, and $\rho'$
	\State $\mathbf{Z} \leftarrow \mathbf{E} \leftarrow f(\mathbf{A}; \phi)$ \Comment Run the unified encoder
	\State $\hat{\mathbf{X}}, \hat{\mathbf{Y}} \leftarrow g_x (\mathbf{Z}, \mathbf{A}; \theta), g_y (\mathbf{Z}, \mathbf{A}; \rho)$ \Comment Make predictions
	\State $l_{xy} \leftarrow \sum_i l_x(\hat{\mathbf{x}}_i, \mathbf{x}_i) + \sum_j l_y(\hat{\mathbf{y}}_j, \mathbf{y}_j)$ \Comment Equation \eqref{eq:loss-x} to \eqref{eq:loss-x3}
	\State $\mathbf{K} \leftarrow \mathbf{I} - \mathbf{D}^{-1/2} \adj \mathbf{D}^{-1/2}$ \Comment Equation \eqref{eq:info-matrix}
	\State $l_\mathrm{GMRF} \leftarrow \mathrm{tr}(\mathbf{E}^\top \mathbf{K} \mathbf{E}) - \alpha \log|\mathbf{I} + \beta^{-1}\mathbf{E}^\top \mathbf{E}|$  \Comment Equation \eqref{eq:kl-divergence-2}
	\State$\phi', \theta', \rho' \leftarrow$ Update $\phi, \theta, \rho$ to minimize $l_{xy} + \lambda l_\mathrm{GMRF}$
\end{algorithmic}
\label{alg:overview}
\end{algorithm}

\textbf{Unit normalization.}
A possible limitation of introducing the identity feature matrix is the large size of $\mathbf{W}_1$, which can make the training process unstable.
Thus, we project the latent representations $\mathbf{Z}$ generated from the encoder $f$ into a unit hypersphere by normalizing each vector of node $i$ as $\mathbf{z}_i / \|\mathbf{z}_i \|_2$.
This does not alter the main functionality of making diverse representations of nodes for making high-dimensional features, but improves the stability of training by restricting the output space \cite{Ying2018}.

\subsubsection{Decoder Networks}
We propose two decoder networks $g_x$ and $g_y$ to model $p_\theta(\mathbf{x}_i \mid \mathbf{z}_i)$ and $p_\rho (y_i \mid \mathbf{z}_i)$, respectively.
We assume that latent variables $\mathbf{Z}$ have sufficient information to construct the observed features and labels.
Thus, we minimize the complexity of decoder networks by adopting the simplest linear transformation as $g_x(\mathbf{z}_i) = \mathbf{W}_x \mathbf{z}_i + \mathbf{b}_x$ and $g_y(\mathbf{z}_i) = \mathbf{W}_y \mathbf{z}_i + \mathbf{b}_y$, where {\small $\mathbf{W}_x \in \mathbb{R}^{m \times d}$}, {\small $\mathbf{W}_y \in \mathbb{R}^{c \times d}$}, $\mathbf{b}_x \in \mathbb{R}^m$ and $\mathbf{b}_y \in \mathbb{R}^c$ are learnable weights and biases, $m$ is the number of features, and $c$ is the number of classes.

\subsubsection{Optimization}

We update the parameters of all the networks $f$, $g_x$, and $g_y$ in an end-to-end way.
We rewrite the ELBO of Equation \eqref{eq:elbo-1} as the following objective function to be minimized:
\begin{equation}
	l(\Theta) = \sum_{i \in \mathcal{V}_x} l_x(\mathbf{\hat{x}}_i, \mathbf{x}_i) + \sum_{i \in \mathcal{V}_y} l_y(\hat{\mathbf{y}}_i, \mathbf{y}_i) + \lambda l_\mathrm{GMRF}(\mathbf{Z}, \adj),
\label{eq:obj-function}
\end{equation}
where $l_x$ and $l_y$ are loss terms for features and labels, respectively.
$l_\mathrm{GMRF}$ is our proposed regularizer, whose details are described in Sections \ref{ssec:method-variational-inference} and \ref{ssec:deterministic-modeling} through the process of structured inference.
We use a hyperparameter $\lambda$ for the amount of regularization.

The loss terms $l_x$ and $l_y$ are determined by how we model the distributions $p_\theta(\mathbf{x}_i \mid \mathbf{z}_i)$ and $p_\rho (y_i \mid \mathbf{z}_i)$ following the distribution of true data.
Common distributions for features include Gaussian, Bernoulli, and categorical (or one-hot) distributions:
\begin{equation}
	l_x(\mathbf{\hat{x}}_i, \mathbf{x}_i) = \begin{cases}
		l_{\mathrm{gau}}(\mathbf{\hat{x}}_i, \mathbf{x}_i) & \textrm{if $\mathbf{x}_i$ is continuous} \\
		l_{\mathrm{ber}}(\mathbf{\hat{x}}_i, \mathbf{x}_i) & \textrm{if $\mathbf{x}_i$ is binary} \\
		l_{\mathrm{cat}}(\mathbf{\hat{x}}_i, \mathbf{x}_i) & \textrm{if $\mathbf{x}_i$ is categorical},
	\end{cases}
	\label{eq:loss-x}
\end{equation}
where the specific loss terms are defined as follows:
\begin{align}
	& l_{\mathrm{gau}}(\mathbf{\hat{x}}_i, \mathbf{x}_i)
		= - {\textstyle \sum_k} ( x_{ik} - \hat{x}_{ik} )^2 \label{eq:loss-x1} \\
	& l_{\mathrm{ber}}(\mathbf{\hat{x}}_i, \mathbf{x}_i)
		= - {\textstyle \sum_k} ( \alpha x_{ik} \log \sigma(\hat{x}_{ik}) \notag \\
	& \quad \quad \quad \quad \quad \quad \quad \quad + (1 - \alpha) (1 - x_{ik}) \log (1 - \sigma(\hat{x}_{ik})) ) \label{eq:loss-x2} \\
	& l_\mathrm{cat}(\hat{\mathbf{x}}_i, \mathbf{x}_i)
		= - {\textstyle \sum_k} x_{ik} \log \mathrm{softmax} (\hat{x}_{ik}). \label{eq:loss-x3}
\end{align}
$\hat{\mathbf{x}}_i = g_x(\mathbf{z}_i)$ is the output of the feature decoder, and $\sigma$ is the logistic sigmoid function.
We introduce $\alpha$ in Equation \eqref{eq:loss-x2} to balance the effects of zero and nonzero entries of true features based on their occurrences \cite{Chen2020}; $\alpha$ is the ratio of zero entries in the observed feature matrix.
For the output $\mathbf{y}_i = g_y(\mathbf{z}_i)$ of the label decoder, we use the categorical loss, i.e., $l_y = l_\mathrm{cat}$, due to the property of labels.

Algorithm \ref{alg:overview} summarizes the training process of \method.
It makes latent variables and predictions in lines 1 and 2, respectively, and computes the error between predictions and observations in line 3.
Then, it computes our regularizer function in lines 4 and 5, whose information is described in the following subsections, to update the parameters of all three networks in an end-to-end way.

\subsection{Structured Variational Inference}
\label{ssec:method-variational-inference}

Previous works utilizing variational inference \cite{Kingma2014, Kipf2016} assume the prior of latent variables as a multivariate Gaussian distribution with identity covariance matrices, and run inference independently for each variable.
%This assumption is inappropriate in our case due to the following reasons.
%First, each variable $Z_i$ should be identifiable in our case, since our objective is to estimate specific $\mathbf{x}_i$ instead of generating any plausible samples.
This assumption is inappropriate in our case, since the correlations between variables, represented as a graph, are the main evidence in our graph-based learning.

We thus model the prior distribution $p(\zvar \mid \adj)$ of Equation \eqref{eq:elbo-1} as Gaussian Markov random field (GMRF) to incorporate the graph structure in the probabilistic modeling of variables.
Specifically, we model $p(\zvar \mid \adj)$ as GMRF $\mathcal{N}(\mathbf{0}, \mathbf{K}^{-1})$ with parameters $\mathbf{h} = \mathbf{0}$ and $\mathbf{K}$.
We make the information matrix $\mathbf{K}$ from $\adj$ as a graph Laplacian matrix with symmetric normalization \cite{Zhang2015}:
\begin{equation}
	\mathbf{K} = \mathbf{I} - \mathbf{D}^{-1/2} \adj \mathbf{D}^{-1/2},
\label{eq:info-matrix}	
\end{equation}
where $\mathbf{I}$ is the identity matrix, and $\mathbf{D}$ is the degree matrix such that $D_{ii} = \sum_j A_{ij}$.
The resulting $\mathbf{K}$ preserves the structural information of the graph $G$ as a positive-semidefinite matrix that satisfies the constraint of GMRF; the nonzero entries of $\mathbf{K}$ except the diagonal ones correspond to those of $\adj$.
Note that $\mathbf{K}$ is a constant, since it represents the fixed prior distribution of variables.

We also model our target distribution $\zdist$ as a multivariate Gaussian distribution $\mathcal{N}(\mathbf{U}, \Sigma)$, where $\mathbf{U}$ and $\Sigma$ are the mean and covariance matrices of size $n \times d$ and $n \times n$, respectively.
We assume that all $d$ elements at each node share the same covariance matrix.
$\mathbf{U}$ and $\Sigma$ are generated from encoder functions $f_\mu$ and $f_\sigma$, respectively, which contain the set $\phi$ of learnable parameters.

Given the Gaussian modelings of $\zdist$ and $\prior$, the KL divergence is formulated as follows:
\begin{multline}
	\kld{q_\phi (\zvar \mid \xall, \yall, \adj)}{p(\zvar \mid \adj)} \\
	= 0.5 ( \mathrm{tr}(\mathbf{U}^\top \mathbf{K} \mathbf{U}) + d(\mathrm{tr}(\mathbf{K}\Sigma) - \log|\Sigma|)) + C,
	\label{eq:kl-divergence}
\end{multline}
where $C$ is a constant related to $\mathbf{K}$ and $|\mathcal{V}|$.
The goal of minimizing the KL divergence is to update $\phi$ of encoder functions to make $q_\phi$ similar to $p(\zvar \mid \adj)$ as a regularizer of latent variables.

The computational bottleneck of Equation \eqref{eq:kl-divergence} is $\log |\Sigma|$, whose computation is $O(n^3)$ \cite{Han2015}.
Thus, we decompose the covariance as $\Sigma = \beta \mathbf{I} + \mathbf{V} \mathbf{V}^\top$ with a rectangular matrix $\mathbf{V} \in \mathbb{R}^{n \times r}$, where $\beta$ and $r$ are hyperparameters such that $r \ll n$ \cite{Tomczak2020}.
As a result, $\log |\Sigma|$ is computed efficiently by the matrix determinant lemma \cite{Harville1998}:
\begin{equation}
	\log |\Sigma| = \log|\mathbf{I}_r + \beta^{-1} \mathbf{V}^\top \mathbf{V}| + \log |\beta\mathbf{I}_n|,
\label{eq:det-lemma}
\end{equation}
where $\mathbf{I}_r$ and $\mathbf{I}_n$ are the identity matrices of sizes $r \times r$ and $n \times n$, respectively.
The computation of Equation \eqref{eq:det-lemma} is $O(r^2n + r^3)$, which is tractable even in graphs with a large number of nodes.

For each inference, we sample $\mathbf{Z}$ randomly from $q_\phi$ based on $\mathbf{U}$ and $\mathbf{V}$ generated from $f_\mu$ and $f_\sigma$, respectively.
Since the gradient-based update is not possible with the direct sampling of $\zvar$, we use the reparametrization trick of variational autoencoders \cite{Kingma2014, Tomczak2020}:
\begin{equation}
	\mathbf{Z} = \mathbf{U} + \sqrt{\beta} \mathbf{M}_1 + \mathbf{V}\mathbf{M}_2,
	\label{eq:var-sampling}
\end{equation}
where $\mathbf{M}_1 \in \mathbb{R}^{n \times d}$ and $\mathbf{M}_2 \in \mathbb{R}^{r \times d}$ are matrices of standard normal variables, which are sampled randomly at each time to simulate the sampling of $\zvar$ while supporting the backpropagation.
The detailed process of inference is described in Appendix \ref{appendix:algorithm}.

We verify that the variables $\mathbf{Z}$ sampled from Equation \eqref{eq:var-sampling} follow the target distribution $\mathcal{N}(\mathbf{U}, \Sigma)$ by Lemmas \ref{lemma:vae-1} and \ref{lemma:vae-2}.

\begin{lemma}
	Let $\mathbf{z}_i$ be a latent variable sampled from Equation \eqref{eq:var-sampling} for node $i$, and $\mathbf{u}_i$ be the $i$-th row of $\mathbf{U}$.
	Then, $\mathbb{E}[\mathbf{z}_i] = \mathbf{u}_i$.
	\label{lemma:vae-1}
\end{lemma}

\begin{proof}
	\vspace{-1mm}
	See Appendix \ref{appendix:proof-vae-1}.
	\vspace{-1mm}
\end{proof}

\begin{lemma}
	Assume that the size $d$ of latent variables is one.
	Let $z_i$ and $z_j$ be latent variables sampled from Equation \eqref{eq:var-sampling} for nodes $i$ and $j$, respectively.
	Then, $\mathbb{E}[(z_i - \mathbb{E}[z_i])(z_j - \mathbb{E}[z_j])] = \Sigma_{ij}$.
	\label{lemma:vae-2}
\end{lemma}

\begin{proof}
	\vspace{-1mm}
	See Appendix \ref{appendix:proof-vae-2}.
	\vspace{-1mm}
\end{proof}

\begin{table*}
	\setlength\tabcolsep{4.9pt}
	\centering
	\caption{%
		Evaluation of \method and baseline approaches for missing feature estimation with respect to (top) recall and (bottom) nDCG.
		The best is in bold, and the second best is underlined.
		Our \method outperforms all baselines in most cases.
	}
	\vspace{-1mm}
	\small
	\begin{tabular}{c|l|ccc|ccc|ccc|ccc|ccc}
		\toprule
		\multirow{2}{*}{\textbf{Metric}}
			& \multirow{2}{*}{\textbf{Model}}
			& \multicolumn{3}{c|}{\textbf{Cora}}
			& \multicolumn{3}{c|}{\textbf{Citeseer}}
			& \multicolumn{3}{c|}{\textbf{Computers}}
			& \multicolumn{3}{c|}{\textbf{Photo}}
			& \multicolumn{3}{c}{\textbf{Steam}} \\
		& & $@10$ & $@20$ & $@50$
			& $@10$ & $@20$ & $@50$
			& $@10$ & $@20$ & $@50$
			& $@10$ & $@20$ & $@50$
			& $@3$ & $@5$ & $@10$ \\
		\midrule
		\midrule
		\multirow{8}{*}{Recall} & NeighAgg
			& .0906 & .1413 & .1961
			& .0511 & .0908 & .1501
			& .0321 & .0593 	& .1306
			& .0329 & .0616 & .1361
			& .0603 & .0881 & .1446 \\
		& VAE
			& .0887 & .1228 & .2116
			& .0382 & .0668 & .1296
			& .0255 & .0502 	& .1196
			& .0276 & .0538 & .1279
			& .0564 & .0820 & .1251 \\
		& GNN*
			& .1350 & .1812 & .2972
			& .0620 & .1097 & .2058
			& .0273 & .0533 	& .1278
			& .0295 & .0573 & .1324
			& .2395 & .3431 & .4575 \\
		& GraphRNA
			& .1395 & .2043 & .3142
			& .0777 & .1272 & .2271
			& .0386 & .0690 	& .1465
			& .0390 & .0703 & .1508
			& .2490 & .3208 & .4372 \\
		& ARWMF
			& .1291 & .1813 & .2960
			& .0552 & .1015 & .1952
			& .0280 & .0544 	& .1289
			& .0294 & .0568 & .1327
			& .2104 & .3201 & .4512 \\
%		& SAT-SAGE
%			& .1356 & .1981 & .3165
%			& .0704 & .1163 & .2174
%			& .0419 & .0738 	& .1562
%			& \textbf{.0483} & \underline{.0766} & .1601
%			& .2518 & .3470 & .4845 \\
		& SAT
			& \underline{.1653} & \underline{.2345} & \underline{.3612}
			& \underline{.0811} & \underline{.1349} & \underline{.2431}
			& \underline{.0421} & \underline{.0746} & \underline{.1577}
			& \underline{.0427} & \underline{.0765} & \underline{.1635}
			& \underline{.2536} & \textbf{.3620} & \underline{.4965} \\
		\cmidrule{2-17}
		& \textbf{\method}
			& \textbf{.1718} & \textbf{.2486} & \textbf{.3814}
			& \textbf{.0943} & \textbf{.1539} & \textbf{.2782}
			& \textbf{.0437} & \textbf{.0769} & \textbf{.1602}
			& \textbf{.0446} & \textbf{.0798} & \textbf{.1670}
			& \textbf{.2565} & \textbf{.3620} & \textbf{.4996} \\
		\midrule
		\midrule
		\multirow{8}{*}{nDCG} & NeighAgg
			& .1217 & .1548 & .1850
			& .0823 & .1155 & .1560
			& .0788 & .1156 	& .1923
			& .0813 & .1196 & .1998
			& .0955 & .1204 & .1620 \\
		& VAE
			& .1224 & .1452 & .1924
			& .0601 & .0839 & .1251
			& .0632 & .0970 	& .1721
			& .0675 & .1031 & .1830
			& .0902 & .1133 & .1437 \\
		& GNN*
			& .1791 & .2099 & .2711
			& .1026 & .1423 & .2049
			& .0673 & .1028 	& .1830
			& .0712 & .1083 & .1896
			& .3366 & .4138 & .4912 \\
		& GraphRNA
			& .1934 & .2362 & .2938
			& .1291 & .1703 & .2358
			& .0931 & .1333 	& .2155
			& .0959 & .1377 & .2232
			& .3437 & .4023 & .4755 \\
		& ARWMF
			& .1824 & .2182 & .2776
			& .0859 & .1245 & .1858
			& .0694 & .1053 	& .1851
			& .0727 & .1098 & .1915
			& .3066 & .3877 & .4704 \\
%		& SAT-SAGE
%			& .1905 & .2320 & .2947
%			& .1179 & .1563 & .2227
%			& \underline{.1030} & .1457 	& .2333
%			& \underline{.1082} & .1475 & .2402
%			& .3529 & .4271 & .5133 \\
		& SAT
			& \underline{.2250} & \underline{.2723} & \underline{.3394}
			& \underline{.1385} & \underline{.1834} & \underline{.2545}
			& \underline{.1030} & \underline{.1463} & \underline{.2346}
			& \underline{.1047} & \underline{.1498} & \underline{.2421}
			& \textbf{.3585} & \textbf{.4400} & \underline{.5272} \\
		\cmidrule{2-17}
		& \textbf{\method}
			& \textbf{.2381} & \textbf{.2894} & \textbf{.3601}
			& \textbf{.1579} & \textbf{.2076} & \textbf{.2892}
			& \textbf{.1068} & \textbf{.1509} & \textbf{.2397}
			& \textbf{.1084} & \textbf{.1549} & \textbf{.2472}
			& \underline{.3567} & \underline{.4391} & \textbf{.5299} \\
		\bottomrule
	\end{tabular}
	\vspace{-2mm}

	\label{table:feature-estimation}
\end{table*}

\subsection{Unified Deterministic Modeling}
\label{ssec:deterministic-modeling}

The reparametrization trick of variational inference requires us to sample different $\mathbf{Z}$ at each inference to approximate the expectation term in Equation~\eqref{eq:elbo-3}.
However, this sampling process makes the training unstable, considering the characteristics of our feature estimation problem where 1) the inference is done for all nodes at once, not for each node independently, and 2) only a part of target variables have meaningful observations.
Even a small perturbation for each node can result in a catastrophic change of the prediction, since we consider the correlations between nodes in $\mathcal{N}(\mathbf{U}, \Sigma)$.

\begin{figure}
	\centering
	\includegraphics[width=0.44\textwidth]{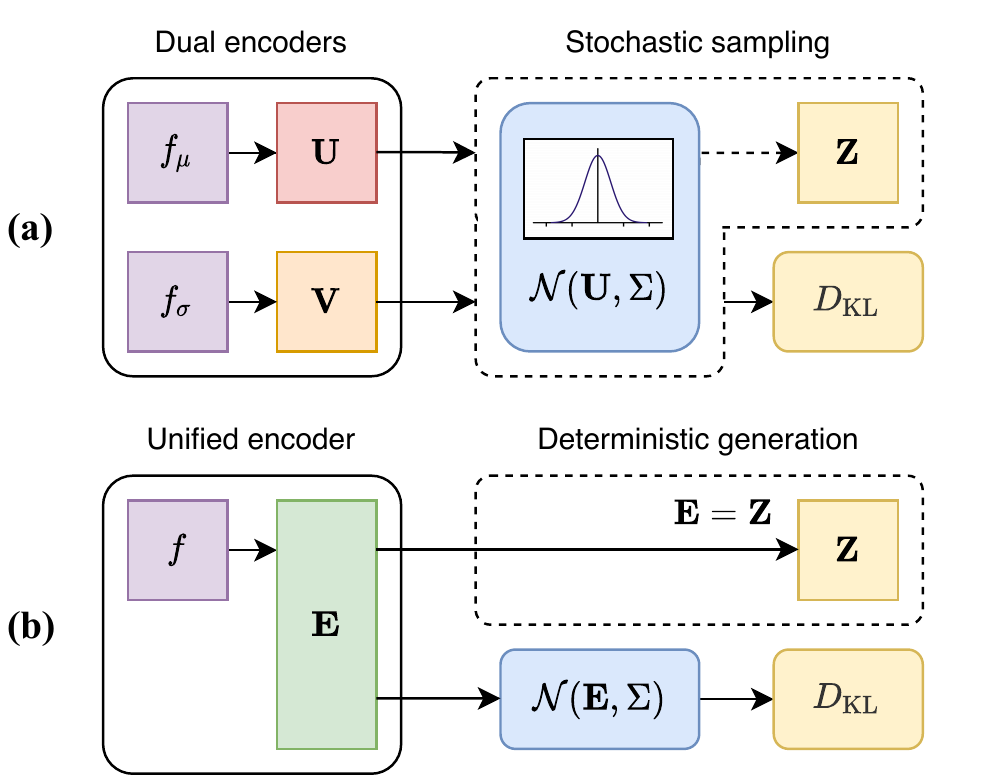}
	\caption{%
		Comparison between the encoder structures of the (a) basic parameterization and (b) unified modeling.
		We use a single encoder $f$ to deterministically generate $\zvar$ while utilizing the strong regularization of the KL divergence.
	}
	\label{fig:unification}
\end{figure}

We propose two ideas for improving the basic parameterization.
First, we unify the parameter matrices $\mathbf{U}$ and $\mathbf{V}$ as a single matrix $\mathbf{E}$, and generate it from an encoder function $f$.
This is based on the observation that $\mathbf{U}$ and $\mathbf{V}$ have similar roles of representing target nodes as low-dimensional vectors based on the graphical structure.
Second, we change the stochastic sampling of $\mathbf{Z}$ from $\mathcal{N}(\mathbf{E}, \Sigma)$ into a deterministic process that returns $\mathbf{E}$ at every inference, which has the largest probability in the distribution of $\mathbf{Z}$.
This improves the stability of inference, while still allowing us to regularize the distribution of $\mathbf{Z}$ with the KL divergence.
Figure \ref{fig:unification} depicts the difference between the basic parameterization and the unified modeling.

This unified modeling makes the KL divergence of Equation~\eqref{eq:kl-divergence} into a general regularizer function that works with deterministic inference of node representations.
First, we show in Lemma \ref{lemma:equivalence} that the first two terms of the right hand side of Equation~\eqref{eq:kl-divergence} become equivalent as we assume $\mathbf{E} = \mathbf{U} = \mathbf{V}$ by the unified modeling.

\begin{lemma}
	Let $\mathbf{K} \in \mathbb{R}^{n \times n}$, $\mathbf{E} \in \mathbb{R}^{n \times d}$, and $\Sigma = \beta \mathbf{I} + \mathbf{E} \mathbf{E}^\top$.
	Then, $\mathrm{tr}(\mathbf{K} \Sigma) = \mathrm{tr}(\mathbf{E}^\top \mathbf{K} \mathbf{E}) + C$, where $C$ is a constant unrelated to $E$.
\label{lemma:equivalence}
\end{lemma}

\begin{proof}
	\vspace{-1mm}
	See Appendix \ref{appendix:proof-equivalence}.
	\vspace{-1mm}
\end{proof}

Then, we propose our regularizer function used in Algorithm \ref{alg:overview} by rewriting the KL divergence of Equation \eqref{eq:kl-divergence} as follows:
\begin{equation}
	l_\mathrm{GMRF}(\mathbf{E}, \mathbf{A}) = \mathrm{tr}(\mathbf{E}^\top \mathbf{K} \mathbf{E}) - \alpha \log|\mathbf{I} + \beta^{-1}\mathbf{E}^\top \mathbf{E}|,
\label{eq:kl-divergence-2}
\end{equation}
where $\alpha > 0$ is a hyperparameter that controls the effect of the log determinant term.
We set $\alpha = 1/2$ is all of our experiments. 
%Then, the KL divergence of Equation \eqref{eq:kl-divergence} is rewritten as
%\begin{equation}
%	l_\mathrm{GMRF}(\mathbf{E}, \mathbf{A}) = \mathrm{tr}(\mathbf{E}^\top \mathbf{K} \mathbf{E}) - 0.5 \log|\mathbf{I} + \beta^{-1}\mathbf{E}^\top \mathbf{E}| + C,
%\label{eq:kl-divergence-2}
%\end{equation}
%where $C$ is the same constant as in Equation \eqref{eq:kl-divergence}.
%This is our proposed regularizer to train \method in Algorithm \ref{alg:overview}.

The first term of $l_\mathrm{GMRF}$ is called the graph Laplacian regularizer and has been widely used in graph learning \cite{Ando06, Pang2017}.
Its minimization makes adjacent nodes have similar representations in $\mathbf{E}$, and the symmetric normalization of $\mathbf{K}$ alleviates the effect of node degrees in the regularization.
%\begin{equation}
%	\mathrm{tr}(\mathbf{E}^\top \mathbf{K} \mathbf{E})
%	= \sum_{(i, j) \in \mathcal{E}} \Bigl\| \frac{\mathbf{e}_i}{\sqrt{d_i}} - \frac{\mathbf{e}_j}{\sqrt{d_j}}\Bigr\|^2_2,
%\label{eq:dot-product-2}
%\end{equation}
%where $\mathbf{e}_i$ and $\mathbf{e}_j$ are the $i$-th and the $j$-th row of $\mathbf{E}$, and $d_i$ and $d_j$ are the degrees of nodes $i$ and $j$, respectively.
%The minimization of Equation \eqref{eq:dot-product-2} makes adjacent nodes have similar representations in $\mathbf{E}$, and the symmetric normalization of $\mathbf{K}$ alleviates the effect of different node degrees in the regularization.
The second term of $l_\mathrm{GMRF}$ can be considered as measuring the amount of space occupied by $\mathbf{E}$.
In other words, its maximization makes $\mathbf{e}_1, \cdots, \mathbf{e}_n$ distributed sparsely, alleviating the effect of $\mathrm{tr}(\mathbf{E}^\top \mathbf{K} \mathbf{E})$ that squeezes the embeddings into a small space.
The hyperparameter $\beta$ controls the balance between the two terms having opposite goals; the second term is ignored if $\beta = \infty$, which means that the target nodes have no correlations.

\subsection{Complexity Analysis}
\label{subsec:analysis}

We analyze the time and space complexities of \method, assuming a graph convolutional network having two layers as the encoder function $f$.
We define a space complexity as the amount of space required to store intermediate data during each inference.
Let $d$, $m$, and $c$ be the size of latent variables, the number of node features, and the number of labels, respectively.

\begin{lemma}
	Given a graph $G = (\mathcal{V}, \mathcal{E})$, the time complexity of \method is $O((d^2 + md + cd)|\mathcal{V}| + d|\mathcal{E}|)$ for each inference.
\label{lemma:time-complexity}
\end{lemma}

\begin{proof}
	\vspace{-1mm}
	See Appendix \ref{appendix:proof-time-complexity}.
	\vspace{-1mm}
\end{proof}

\begin{lemma}
	Given a graph $G = (\mathcal{V}, \mathcal{E})$, the space complexity of \method is $O((d + m + c)|\mathcal{V}| + |\mathcal{E}| + d^2 + md + cd)$ for each inference.
\label{lemma:space-complexity}
\end{lemma}

\begin{proof}
	\vspace{-1mm}
	See Appendix \ref{appendix:proof-space-complexity}.
	\vspace{-1mm}
\end{proof}

Lemmas \ref{lemma:time-complexity} and \ref{lemma:space-complexity} show that \method is an efficient method whose complexity is linear with both the numbers of nodes and edges of the graph.
The GMRF regularizer does not affect the inference of \method, because it is used only at the training time.
Still, the time and space complexities of the GMRF loss $l_\mathrm{GMRF}$ of Equation \eqref{eq:kl-divergence-2} are $O(d^2|\mathcal{V}| + d|\mathcal{E}| + d^3)$ and $O(d|\mathcal{V}| + |\mathcal{E}| + d^2)$, respectively, which are linear with both the numbers of nodes and edges.

\begin{table}
	\centering
	\caption{%
		Summary of datasets.
	}
	\vspace{-1mm}

	\small
	\begin{tabular}{l|c|rrrr}
		\toprule
		\textbf{Dataset}
			& \textbf{Type}
			& \textbf{Nodes}
			& \textbf{Edges}
			& \textbf{Feat.}
			& \textbf{Classes} \\
		\midrule
		Cora\textsuperscript{1}
			& Binary & 2,708 & 5,429 & 1,433 & 7 \\
		Citeseer\textsuperscript{1}
			& Binary & 3,327 & 4,732 & 3,703 & 6 \\
		Photo\textsuperscript{2}
			& Binary & 7,650 & 119,081 & 745 & 8 \\
		Computers\textsuperscript{2}
			& Binary & 13,752 & 245,861 & 767 & 10 \\
		Steam\textsuperscript{3}
			& Binary & 9,944 & 266,981 & 352 & 1 \\
		\midrule
		Pubmed\textsuperscript{1}
			& Continuous & 19,717 & 44,324 & 500 & 3 \\
		Coauthor\textsuperscript{2}
			& Continuous & 18,333 & 81,894 & 6,805 & 15 \\
		Arxiv\textsuperscript{4}
			& Continuous & 169,343 & 1,157,799 & 128 & 40 \\
		\bottomrule
	\end{tabular}

	\begin{flushleft} \footnotesize
		\ \ \ \ \textsuperscript{1} \url{https://github.com/kimiyoung/planetoid} \\
		\ \ \ \ \textsuperscript{2} \url{https://github.com/shchur/gnn-benchmark} \\
		\ \ \ \ \textsuperscript{3} \url{https://github.com/xuChenSJTU/SAT-master-online} \\
		\ \ \ \ \textsuperscript{4} \url{https://ogb.stanford.edu/docs/nodeprop/}
	\end{flushleft}

%	\vspace{-2mm}
	\label{table:datasets}
\end{table}

\section{Experiments}
\label{sec:experiments}

We perform experiments to answer the following questions:
\begin{itemize}
	\item[Q1.] \textbf{Feature estimation (Section \ref{subsec:feature-estimation}).}
		Does \method show higher accuracy in feature estimation than those of baselines?
	\item[Q2.] \textbf{Node classification (Section \ref{subsec:node-classification}).}
		Are the features generated by \method meaningful for node classification?
	\item[Q3.] \textbf{Effect of observed labels (Section \ref{subsec:with-labels}).}
		Does the observation of labels help generating more accurate features?
	\item[Q4.] \textbf{Scalability (Section \ref{ssec:scalability}).}
		How does the computational time of \method increase with the number of edges?
	\item[Q5.] \textbf{Ablation study (Section \ref{subsec:ablation-study}).}
		How does the performance of \method for feature estimation change by the GMRF regularizer and the deterministic modeling of inference?
\end{itemize}

\begin{table}
	\setlength\tabcolsep{4pt}

	\centering
	\caption{
		Evaluation for missing feature estimation on continuous features.
		The best is in bold, and the second best is underlined.
		RMSE is lower the better, while CORR is higher the better.
		``o.o.m.'' refers to an out-of-memory error.
	}
	\vspace{-1mm}

	\small
	\begin{tabular}{l|rr|rr|rr}
		\toprule
		\multirow{2}{*}{\textbf{Model}} &
			\multicolumn{2}{c|}{\textbf{Pubmed}} &
			\multicolumn{2}{c|}{\textbf{Coauthor}} &
			\multicolumn{2}{c}{\textbf{Arxiv}} \\
		& \textbf{RMSE} & \textbf{CORR} &
			\textbf{RMSE} & \textbf{CORR} &
			\textbf{RMSE} & \textbf{CORR} \\
		\midrule
		NeighAgg
			& 0.0186
			& -0.2133
			& 0.0952
			& -0.2279
			& 0.1291
			& -0.4943 \\
		VAE
			& 0.0170
			& -0.0236
			& 0.0863
			& -0.0237
			& 0.1091
			& -0.4773 \\
		GNN*
			& 0.0168
			& -0.0010
			& 0.0850
			& 0.0179
			& 0.1091
			& 0.0283 \\
		GraphRNA
			& 0.0172
			& -0.0352
			& 0.0897
			& -0.1052
			& 0.1131
			& -0.0419 \\
		ARWMF
			& \underline{0.0165}
			& \underline{0.0434}
			& 0.0827
			& 0.0710
			& o.o.m.
			& o.o.m. \\
		SAT
			& \underline{0.0165}
			& 0.0378
			& \underline{0.0820}
			& \underline{0.0958}
			& \underline{0.1055}
			& \underline{0.0868} \\
		\midrule
		\textbf{\method}
			& \textbf{0.0158}
			& \textbf{0.1169}
			& \textbf{0.0798}
			& \textbf{0.1488}
			& \textbf{0.1005}
			& \textbf{0.1666} \\
		\bottomrule
	\end{tabular}
	\label{table:continuous-feature-errors}
\end{table}

\begin{table*}
	\setlength\tabcolsep{6pt}
	\centering
	\caption{
		Comparison between \method and baselines by node classification accuracy, where each classifier is trained with the generated features.
		\method outperforms all baseline methods in most cases.
		``o.o.m.'' refers to an out-of-memory error.
	}
	\vspace{-1mm}

	\small
	\begin{tabular}{l|cc|cc|cc|cc|cc|cc|cc}
		\toprule
		\multirow{2}{*}{\textbf{Model}} &
			\multicolumn{2}{c|}{\textbf{Cora}} &
			\multicolumn{2}{c|}{\textbf{Citeseer}} &
			\multicolumn{2}{c|}{\textbf{Computers}} &
			\multicolumn{2}{c|}{\textbf{Photo}} &
			\multicolumn{2}{c|}{\textbf{Pubmed}} &
			\multicolumn{2}{c|}{\textbf{Coauthor}} &
			\multicolumn{2}{c}{\textbf{Arxiv}} \\
		& \textbf{MLP} & \textbf{GCN} &
			\textbf{MLP} & \textbf{GCN} &
			\textbf{MLP} & \textbf{GCN} &
			\textbf{MLP} & \textbf{GCN} &
			\textbf{MLP} & \textbf{GCN} &
			\textbf{MLP} & \textbf{GCN} &
			\textbf{MLP} & \textbf{GCN} \\
		\midrule
		NeighAgg
			& .6248
			& .6494
			& .5539
			& .5413
			& \underline{.8365}
			& .8715
			& .8846
			& .9010
			& .5150
			& .6564
			& .7562
			& .8031
			& \underline{.3979}
			& \underline{.6493} \\
		VAE
			& .2826
			& .3011
			& .2551
			& .2663
			& .3747
			& .4023
			& .2598
			& .3781
			& .4008
			& .4007
			& .2317
			& .2335
			& .1633
			& .1965 \\
		GNN*
			& .4852
			& .5779
			& .3933
			& .4278
			& .3747
			& .4034
			& .2683
			& .3789
			& .4013
			& .4203
			& .2317
			& .2335
			& .2607
			& .4721 \\
		GraphRNA
			& .7581
			& .8198
			& .6320
			& .6394
			& .6968
			& .8650
			& .8407
			& .9207
			& .6035
			& \underline{.8172}
			& \underline{.7710}
			& \underline{.8851}
			& .1609
			& .1859 \\
		ARWMF
			& .7769
			& .8205
			& .2267
			& .2764
			& .5608
			& .7400
			& .4675
			& .6146
			& \textbf{.6180}
			& .8089
			& .2320
			& .8347
			& o.o.m.
			& o.o.m. \\
		SAT
			& \underline{.7937}
			& \textbf{.8579}
			& \underline{.6475}
			& \underline{.6767}
			& .8201
			& \underline{.8766}
			& \underline{.8976}
			& \textbf{.9260}
			& .4618
			& .7439
			& .7672
			& .8402
			& .3144
			& .5677 \\
		\midrule
		\textbf{\method (proposed)}
			& \textbf{.8431}
			& \underline{.8490}
			& \textbf{.6774}
			& \textbf{.6844}
			& \textbf{.8450}
			& \textbf{.8889}
			& \textbf{.9021}
			& \underline{.9253}
			& \underline{.6178}
			& \textbf{.8315}
			& \textbf{.8805}
			& \textbf{.9023}
			& \textbf{.4394}
			& \textbf{.6644} \\
		\bottomrule
	\end{tabular}
	\vspace{-2mm}

	\label{table:classification-accuracy}
\end{table*}

\subsection{Experimental Setup}
\label{ssec:exp-setup}

We introduce our experimental setup including datasets, baseline methods, evaluation metrics, and training processes.

\textbf{Datasets.}
We use graph datasets summarized in Table \ref{table:datasets}, which were used in previous works \cite{McAuley2015, Yang2016, Shchur2018, Chen2020}.
Node features in Cora, Citeseer, Photo, Computers, and Steam are zero-one binary vectors, while those in Pubmed, Coauthor, and ArXiv are continuous.
Each node has a single discrete label.
All nodes in Steam have the same class, and thus the dataset is not used for node classification.

\textbf{Baselines.}
We compare \method with existing models for feature estimation.
NeighAgg \cite{Simsek2008} is a simple approach that aggregates the features of neighboring nodes through mean pooling.
VAE \cite{Kingma2014} is a generative model that learns latent representations of examples.
%We make latent representations of test nodes that contain no initial features by applying NeighAgg to the latent space.
GCN \cite{Kipf2017}, GraphSAGE \cite{Hamilton2017}, and GAT \cite{Velickovic2018} are popular graph neural networks that have been used in various domains.
%We train the models by giving the graph structure as an input, represented as an adjacency matrix, and using the observed features as a target.
We report the best performance among the three models as GNN* for brevity.

GraphRNA \cite{Huang2019} and ARWMF \cite{Chen2019} are recent methods for representation learning, which can be applied for generating features.
SAT \cite{Chen2020} is the state-of-the-art model for missing feature estimation, which trains separate autoencoders with a shared latent space for the features and graphical structure, respectively.
We use GAT and GCN as the backbone network of SAT in datasets with discrete and continuous features, respectively, which are the settings that show the best performance in the original paper \cite{Chen2020}.

\begin{figure}
	\centering
	\includegraphics[trim=0 0 0 0, width=0.29\textwidth]{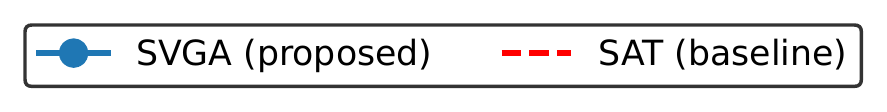}
	\vspace{-1mm}

	\begin{subfigure}{0.232\textwidth}
		\includegraphics[width=\textwidth]{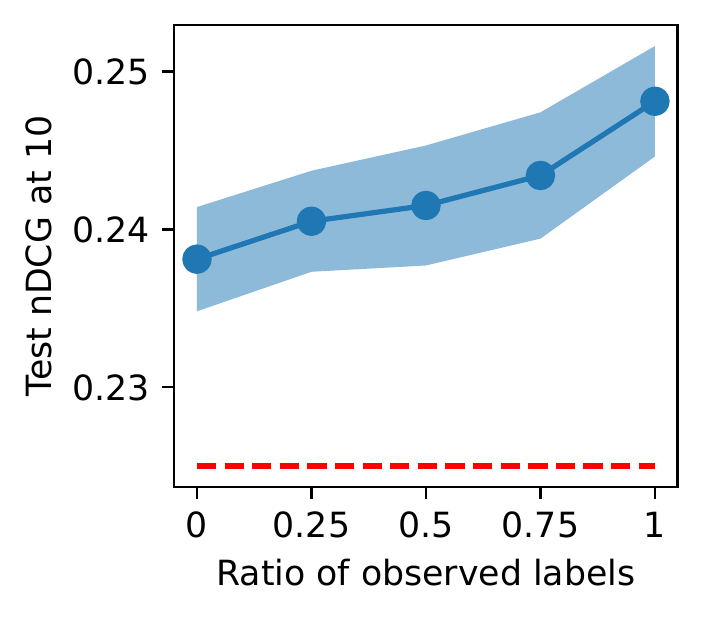}
        \vspace{-1.5\baselineskip}
		\caption{Cora}
	\end{subfigure} \hfill
	\begin{subfigure}{0.239\textwidth}
		\includegraphics[width=\textwidth]{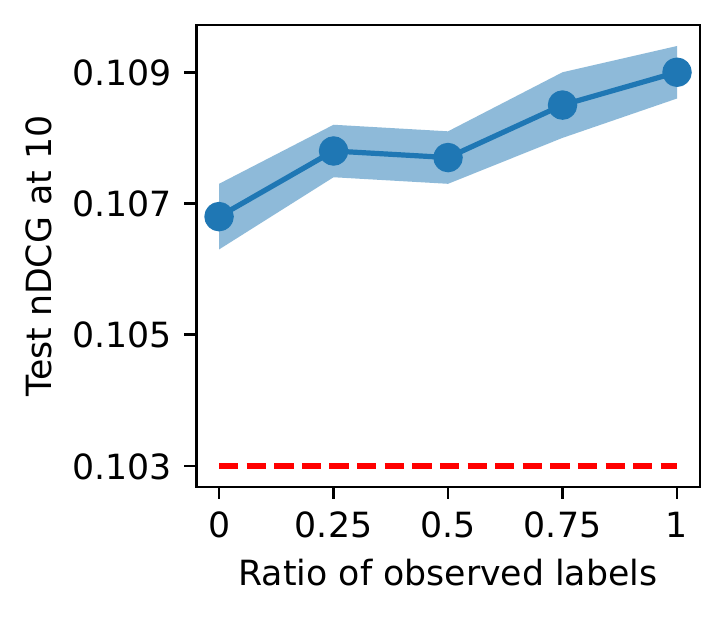}
        \vspace{-1.5\baselineskip}
		\caption{Computers}
	\end{subfigure}
		
	\caption{%
		The accuracy of \method for feature estimation with additional observations of labels.
		We show the average and standard deviation of ten runs.
		\method effectively utilizes the given labels, making more accurate predictions.
	}
	\label{fig:more-labels}
\end{figure}

\textbf{Evaluation metrics.}
We evaluate the performance of feature estimation with four evaluation metrics.
For binary features, we treat each nonzero entry as a target item, considering the task as a ranking problem to find all nonzero entries.
Recall at $k$ measures the ratio of true entries contained in the top $k$ predictions for each node, while nDCG at $k$ measures the overall quality of predicted scores in terms of information retrieval.
We vary $k$ over $\{3, 5, 10\}$ in the Steam dataset and $\{10, 20, 50\}$ in the other datasets, because Steam has fewer features and thus a prediction is generally easier.
For continuous features, we compare the predictions and the true features in an elementwise way with the root mean squared error (RMSE) and the square of the correlation coefficient (CORR).
The definitions of evaluation metrics are in Appendix \ref{appendix:evaluation}.

\textbf{Experimental process.}
We take different processes of experiments for feature estimation and node classification.
For feature estimation, we split all nodes at each dataset into the training, validation, and test sets by the 4:1:5 ratio as in previous work \cite{Chen2020}.
We train each model based on the observed features of training nodes and find the parameters that maximize the validation performance.
We run each experiment ten times and report the average.

For node classification, we take only the test nodes of feature estimation, whose features are generated by our \method or baseline models.
Then, we perform the 5-fold cross-validation in the target nodes, evaluating the quality of generated features with respect to the accuracy of node classification.
We use a multilayer perceptron (MLP) and GCN as classifiers.
For the training and evaluation of GCN, we use the induced subgraph of target nodes.

Even though our \method can utilize observed labels as additional evidence, we do not assume the observation of labels unless otherwise noted.
This is to make a fair comparison between \method and baseline models that assume only the observation of features.
We perform experiments in Section \ref{subsec:with-labels} with observed labels.

\textbf{Hyperparameters.}
The hyperparameter setting of our \method is described in Appendix \ref{appendix:parameters}.
For baselines, we take the experimental results from a previous work \cite{Chen2020} that optimized the hyperparameters for the feature estimation problem on our datasets.

\subsection{Performance on Feature Estimation (Q1)}
\label{subsec:feature-estimation}

Table \ref{table:feature-estimation} compares \method and baseline models for feature estimation.
\method outperforms all baselines with a significant margin in most cases; \method shows up to 16.3\% and 14.0\% higher recall and nDCG, respectively, compared with the best competitors.
The amount of improvement over baselines is the largest in Cora and Citeseer, which are similar citation graphs, and the smallest in Steam.
This is because the citation graphs have high-dimensional features with sparse graph structures, increasing the difficulty of estimation for the baseline methods.
On the other hand, Steam has the smallest number of features, while having the densest structure.

Table \ref{table:continuous-feature-errors} presents the result of feature estimation for continuous features.
\method still outperforms all baselines, and the amount of improvement is similar in all three datasets.
The combination of Tables \ref{table:feature-estimation} and \ref{table:continuous-feature-errors} shows that \method works well with various types of node features, providing stable performance.
ARWMF causes an out-of-memory error in 256GB memory, due to the computation of $\mathbf{A}^n$ of the adjacency matrix $\mathbf{A}$ with large $n\geq5$.

\subsection{Performance on Node Classification (Q2)}
\label{subsec:node-classification}

Table \ref{table:classification-accuracy} shows the accuracy of node classification with two types of classifiers: MLP and GCN.
MLP relies on the generated features for prediction, while GCN utilizes also the graph structure.
\method outperforms all baselines in most cases, making a consistency with the results of feature estimation in Tables~\ref{table:feature-estimation} and \ref{table:continuous-feature-errors};
\method achieves up to 14.2\% and 1.9\% higher accuracy in MLP and GCN, respectively, compared to the best competitors.
The Steam dataset is excluded from Table \ref{table:classification-accuracy}, since it has the same label for all nodes.

\subsection{Effect of Observed Labels (Q3)}
\label{subsec:with-labels}

Figure \ref{fig:more-labels} shows the performance of \method for feature estimation with different ratios of observed labels.
For instance, if the ratio is 0.5, half of all nodes have observed labels: $|\mathcal{V}_y| = 0.5|\mathcal{V}|$.
Note that the experiments for Tables \ref{table:feature-estimation}, \ref{table:continuous-feature-errors}, and \ref{table:classification-accuracy} are done with no labels for a fair comparison with the baseline models; the results of these experiments correspond to the leftmost points in Figure~\ref{fig:more-labels}.
We also report the performance of SAT for comparison.

\method shows higher accuracy with more observations of labels in both datasets, demonstrating its ability to use labels to improve the performance of feature estimation.
Since the parameters need to be optimized to predict both features and labels accurately, the observed labels work as an additional regularizer that guides the training of latent variables to avoid the overfitting.

\subsection{Scalability (Q4)}
\label{ssec:scalability}

\begin{figure}
	\centering

	\begin{minipage}{0.24\textwidth}
		\centering
		\includegraphics[width=\textwidth]{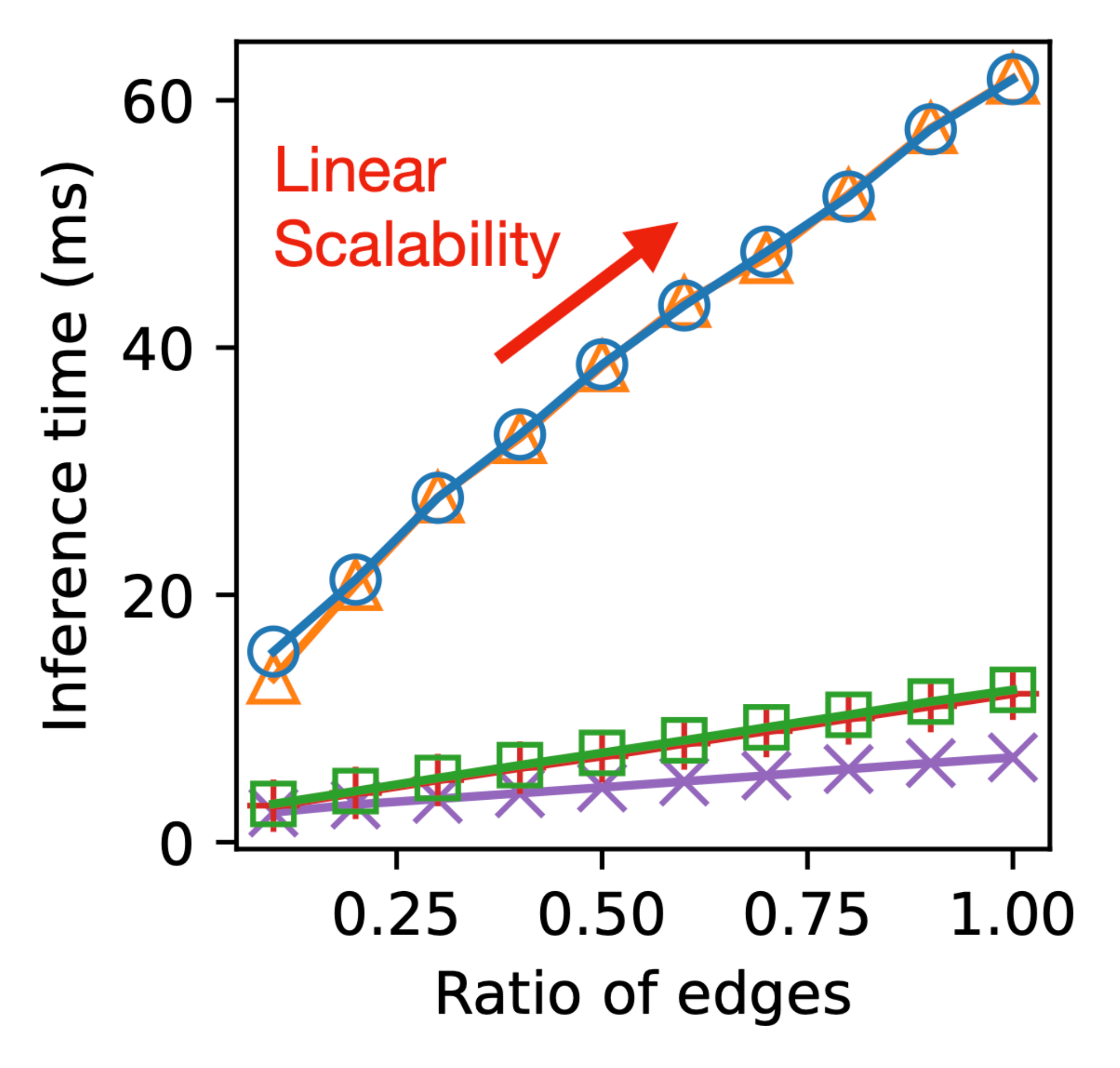}
	\end{minipage} \hspace{-2mm}
	\begin{minipage}{0.125\textwidth}
		\centering
		\vspace{-5mm}
		\includegraphics[width=\textwidth]{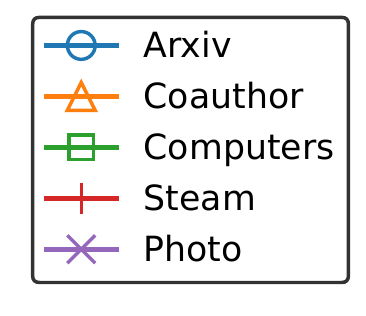}
	\end{minipage}
	
	\vspace{-1mm}
	\caption{
		The inference time of \method in graphs of different sizes.
		We randomly sample nine subgraphs for each dataset.
		\method shows the linear scalability in all datasets.
	}
	\label{fig:scalability}
\end{figure}

Figure \ref{fig:scalability} shows the scalability of \method with respect to the number of edges on the five largest datasets in Table \ref{table:datasets}.
For each dataset, we sample nine random subgraphs of different sizes from $0.1|\mathcal{E}|$ to $0.9|\mathcal{E}|$, where $|\mathcal{E}|$ denotes the number of original edges.
We measure the inference time of \method in each graph ten times and report the average.
The figure shows the linear scalability of \method with the number of edges in all datasets, supporting Lemma~\ref{lemma:time-complexity}.
Arxiv and Coauthor take the longest inference times, as Arxiv and Coauthor have the largest numbers of edges and features, respectively.

\subsection{Ablation Study (Q5)}
\label{subsec:ablation-study}

\begin{figure}
	\centering
	\includegraphics[trim=0 0 0 0, width=0.44\textwidth]{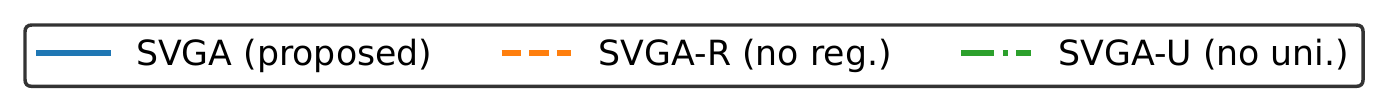}
	\vspace{-1mm}
	
	\begin{subfigure}{0.232\textwidth}
		\includegraphics[width=\textwidth]{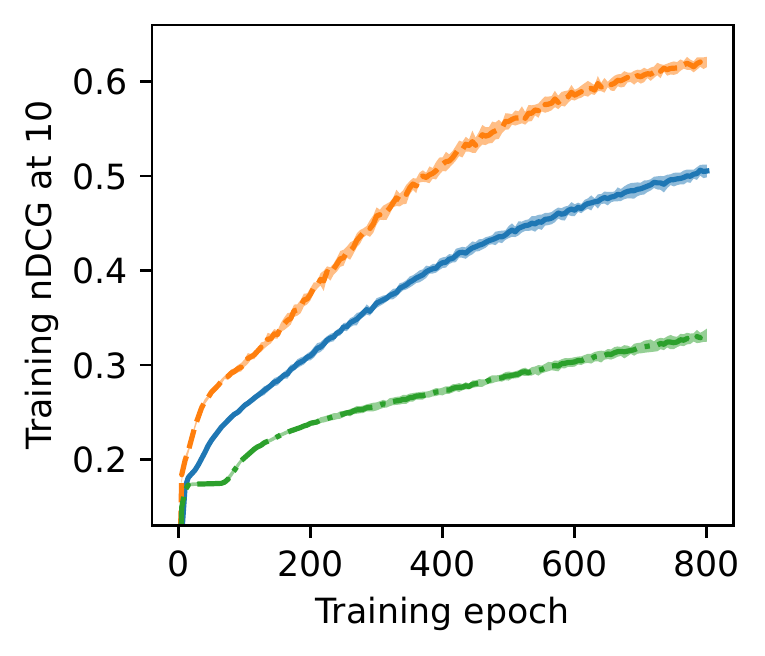}
        \vspace{-1.5\baselineskip}
		\caption{Training}
	\end{subfigure} \hfill
	\begin{subfigure}{0.239\textwidth}
		\includegraphics[width=\textwidth]{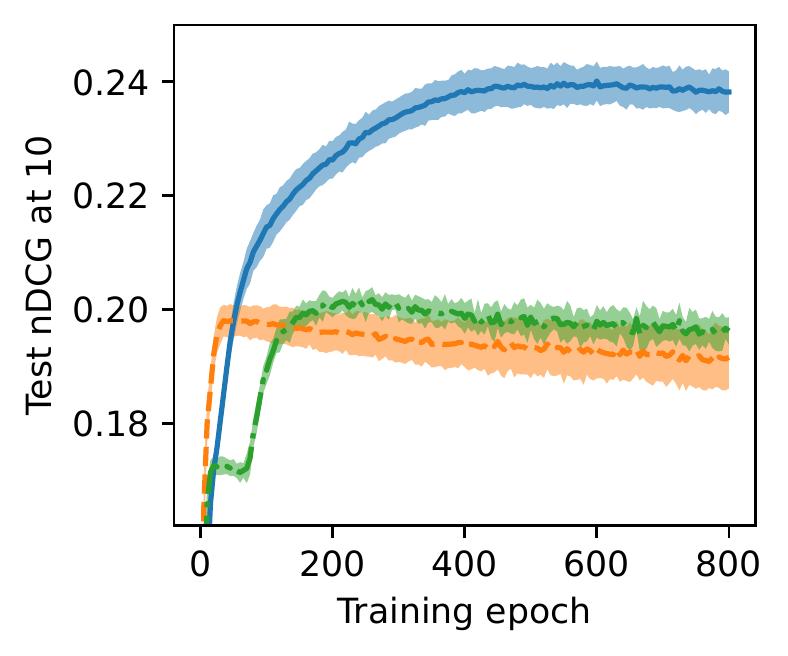}
        \vspace{-1.5\baselineskip}
		\caption{Test}
	\end{subfigure}

	\caption{
		An ablation study of \method on Cora compared with its variants \method-R and \method-U (details in Section \ref{subsec:ablation-study}).
		We show the average and standard deviation of ten runs.
		\method makes the best test accuracy based on our proposed ideas.
	}
	\label{fig:regularization}
\end{figure}

Figure \ref{fig:regularization} shows an ablation study that compares \method with its variants \method-U and \method-R.
\method-U runs stochastic inference described in Section \ref{ssec:method-variational-inference}, without our idea of unified deterministic modeling.
The detailed process of stochastic inference is described also in Algorithm \ref{alg:stochastic} of Appendix \ref{appendix:algorithm}.
\method-R runs the deterministic inference but removes the regularizer term $l_\mathrm{GMRF}$ of Equation~\eqref{eq:kl-divergence-2}; it follows Algorithm \ref{alg:overview} as in \method except for lines 4 and 5.

\method shows the best test accuracy during the training with a stable curve.
The training accuracy is the best with \method-R, since it overfits to training nodes without the regularizer term.
On the other hand, the training accuracy of \method-U is the lowest among the three methods, while its test accuracy becomes similar to that of \method-R at the later epochs.
This is because \method-U fails even at maximizing the training accuracy due to the unstable training.
%The low stability of \method-U is shown also by the large standard deviation of its test accuracy, which increases as training proceeds.
The standard deviation of training accuracy is very small with all three methods, despite their different modelings.

\section{Related Works}
\label{sec:related-works}

%We review related works on \method, which are categorized to graph neural networks, node representation learning, feature estimation, and graph probabilistic modeling.

\textbf{Graph neural networks.}
Graph neural networks (GNN) refer to deep neural networks designed for graph-structured data \cite{Hamilton2017, Velickovic2018, Velickovic2019, Kipf2017, You2020, You2021}.
% Wang2021, Xu2019
Since GNNs require the features of all nodes, one needs to generate artificial features to apply a GNN to a graph with missing features.
\citet{Derr2018} and \citet{Cui2021} generate features from the graph structure.
\citet{Kipf2017} model the missing features as one-hot vectors, while \citet{Zhao2020} leave them as zero vectors and propose a new regularizer function.

Our \method enables a GNN to be applied to graphs with partial observations by estimating missing features.
The main advantage of feature estimation is that the modification of a GNN classifier is not required, regardless of the number of observations given in the original graph.
Previous works that directly deal with partially observed graphs require finding new hyperparameters \cite{Zhao2020} or even making a new weight matrix \cite{Kipf2017} when the number of observations changes, making it difficult to reuse a trained model.

\textbf{Missing feature estimation.}
There are recent works that can be used directly for our feature estimation problem \cite{Huang2019, Chen2019, Chen2020}.
Such methods are adopted as the main competitors in our experiments.
The main advantage of \method over the previous approaches is the strong regularizer that allows us to effectively propagate the partial observations to the entire graph, avoiding the overfitting problem even with large representation power for feature estimation.

GRAPE \cite{You2020} estimates missing features in tabular data by learning a graph between examples and features.
The main difference from our work is that GRAPE assumes partial observations of feature elements, not feature vectors.
In other words, GRAPE cannot be used to estimate the features of nodes that have no partial observations, which is the scenario assumed in our experiments.

\textbf{Node representation learning.}
Unsupervised node representation learning \cite{Hamilton2017b} is to represent each node as a low-dimensional vector that summarizes its properties embedded in the structure and node features \cite{Perozzi2014, Wang2016, Grover2016, Hamilton2017b, Velickovic2019}.
%Traditional methods \cite{Perozzi2014, Wang2016, Grover2016} use only the structural information, while recent ones utilize both the node features and graph structure to maximize performance \cite{Hamilton2017b, Velickovic2019}.
Such methods make embeddings in a latent space, while we aim to learn the representations of nodes in a high-dimensional feature space; the node features generated from our \method are interpretable in the feature domain. %, unlike the embedding vectors.

\textbf{Probabilistic modeling of graphs.}
Previous works model real-world graphs as pairwise Markov random fields with discrete variables and run graphical inference for node classification \cite{Yoo2017, conf/wsdm/JoYK18, Yoo2019b, Yoo2020, Yoo2021}.
Our work can be considered as a generalization of such works into the challenging task of missing feature estimation, which requires us to estimate high-dimensional continuous variables.
%We introduce latent variables to assume the conditional independence between nodes, instead of running graphical inference for marginalization, and propose our deterministic structured variational inference for effective estimation.

\section{Conclusion}
\label{sec:conclusion}

We propose \method (\methodlong), an accurate method for missing feature estimation.
\method estimates high-dimensional features of nodes from a graph with partial observations, and its framework is carefully designed to model the target distributions of structured variational inference.
The main idea of \method is the structural regularizer that assumes the prior of latent variables as Gaussian Markov random field, which considers the graph structure as the main evidence for modeling the correlations between nodes in variational inference.
\method outperforms previous methods for feature estimation and node classification, achieving the state-of-the-art accuracy in benchmark datasets.
Future works include extending the domain of \method into directed or heterogeneous graphs that are common in real-world datasets.

\begin{acks}
{ \small
This work was supported by Institute of Information \& communications Technology Planning \& Evaluation(IITP) grant funded by the Korea government(MSIT) [No.2020-0-00894, Flexible and Efficient Model Compression Method for Various Applications and Environments],
[No.2021-0-01343, Artificial Intelligence Graduate School Program (Seoul National University)],
and
[NO.2021-0-0268, Artificial Intelligence Innovation Hub (Artificial Intelligence Institute, Seoul National University)].
The Institute of Engineering Research at Seoul National University provided research facilities for this work.
The ICT at Seoul National University provides research facilities for this study.
This work was supported by the National Research Foundation of Korea (NRF) grant funded by the Korea government (MSIT) (No. 2021R1C1C1008526).
U Kang is the corresponding author.
}
\end{acks}

\bibliographystyle{ACM-Reference-Format}
\bibliography{paper.bib}

\clearpage
\appendix
%\section{Table of Symbols}
%\label{appendix:symbols}
%
%\blue{Table \ref{table:symbols} summarizes the symbols used frequently in this paper.}
%
%\begin{table}
%\centering
%\caption{Table of symbols.}
%\small
%\begin{tabular}{c|l}
%	\toprule
%	\textbf{Symbol} & \textbf{Description} \\
%	\midrule
%	$G$ & Target undirected graph, consisting of $\mathcal{V}$ and $\mathcal{E}$ \\
%	$\mathcal{V}, \mathcal{E}$ & Sets of nodes and undirected edges, respectively \\
%	$\mathcal{V}_x$ & Subset of nodes having observed features \\
%	$\mathcal{V}_y$ & Subset of nodes having observed labels \\
%%	\midrule
%%	$n$ & The number of nodes: $n = |\mathcal{V}|$ \\
%%	$\mathbf{x}_i, y_i$ & Feature vector and label of node $i \in \mathcal{V}$, respectively \\
%%	$\mathbf{z}_i$ & Latent representation of node $i \in \mathcal{V}$ \\
%%	$y_i$ & Discrete label of node $i \in \mathcal{V}$ \\
%	\midrule
%	$\mathbf{h}, \mathbf{K}$ & Mean vector and precision matrix of GMRF, respectively \\
%	$f_\mu$, $f_\sigma$, $f$ & Encoder networks for latent representations \\
%	$g_x, g_y$ & Decoder networks for features and labels, respectively \\
%	$\mathbf{U}, \mathbf{V}, \mathbf{E}$ & Node embedding matrices for structured inference \\
%	$\mathbf{M}_1, \mathbf{M}_2$ & Standard random matrices for sampling-based inference \\
%	\bottomrule
%\end{tabular}
%\label{table:symbols}
%\end{table}

\section{Proofs of Lemmas}
\label{appendix:proofs-vae}

\subsection{Proof of Lemma \ref{lemma:gmrf}}
\label{appendix:proof-gmrf}

\begin{proof}
	The probability density function of $\mathcal{N}(\mu, \Sigma)$ is
	\begin{equation*}
		f(\mathbf{z}) = C' \exp (-(\mathbf{z} - \mu)^\top \Sigma^{-1} (\mathbf{z} - \mu)),
	\end{equation*}
	where $C' = (2\pi)^{-d/2} |\Sigma|^{-1/2}$ is a constant.
	
	We rewrite $f$ as follows with $\mathbf{K} = \Sigma^{-1}$ and $\mathbf{h} = \mathbf{K} \mu$:
	\begin{align*}
		f(\mathbf{z})
			&= C' \exp (-\mathbf{z}^\top \mathbf{K} \mathbf{z} + 2 \mu^\top \mathbf{K} \mathbf{z} - \mu^\top \mathbf{K} \mu) \\
			&= C'' \exp (-\mathbf{z}^\top \mathbf{K} \mathbf{z} + 2 \mu^\top \mathbf{K} \mathbf{z}) \\
			&= C'' \exp \Bigl( -\sum_i \sum_j z_i K_{ij} z_j + 2 \sum_i h_i z_i \Bigr).
	\end{align*}
	where $C'' = \exp(\mu^\top \mathbf{K} \mu) \cdot C'$ is also a constant.
	
	By the definition of GMRF, $K_{ij} \neq 0$ only if edge $(i, j)$ exists in the given graph $G = (\mathcal{V}, \mathcal{E})$.
	Then, we rewrite $f(\mathbf{z})$ as
	\begin{align*}
		f(\mathbf{z}) = C \exp \Bigl( \sum_{(i, j) \in \mathcal{E}} (- z_i K_{ij} z_j) + \sum_{i \in \mathcal{V}} (-0.5 K_{ii} z_i^2 + h_iz_i) \Bigr),
	\end{align*}
	where $C = \exp(2) \cdot C''$.
	We prove the lemma by substituting $\psi_{ij}(z_{ij})$ and $\psi_i(z_i)$ for the first and second terms, respectively.
\end{proof}

\subsection{Proof of Lemma \ref{lemma:vae-1}}
\label{appendix:proof-vae-1}

\begin{proof}
	The random variables included in Equation \eqref{eq:var-sampling} are $\mathbf{M}_1$ and $\mathbf{M}_2$.
	Since $\mathbf{M}_1$ and $\mathbf{M}_2$ are filled with standard normal values, it is satisfied that $\mathbb{E}[\sqrt{\beta}\mathbf{M}_1] = \mathbf{0}$ and $\mathbb{E}[\mathbf{V} \mathbf{M}_2] = \mathbf{0}$, regardless of the actual values of $\beta$ and $\mathbf{V}$.
	Thus, $\mathbb{E}(\mathbf{Z}) = \mathbb{E}(\mathbf{U}) = \mathbf{U}$.
\end{proof}

\subsection{Proof of Lemma \ref{lemma:vae-2}}
\label{appendix:proof-vae-2}

\begin{proof}
	The following is satisfied for both $k=i$ and $k=j$ based on Lemma \ref{lemma:vae-1}:
	\begin{equation*}
		z_k - \mathbb{E}[z_k] = \sqrt{\beta} m_{1k} + \mathbf{v}_k^\top \mathbf{m}_2,	
	\end{equation*}
	where $\mathbf{v}_k$ and $\mathbf{m}_2$ are $r$-dimensional vectors.

	Then, the covariance between $z_i$ and $z_j$ is given as
	\begin{multline*}
		\mathbb{E}[(z_i - \mathbb{E}[z_i])(z_j - \mathbb{E}[z_j])]
		= \beta \mathbb{E}[m_{1i} m_{1j}] \\
		+ \sqrt{\beta}\mathbf{v}_j^\top \mathbb{E}[m_{1i} \mathbf{m}_2]
		+ \sqrt{\beta} \mathbf{v}_i^\top \mathbb{E}[m_{1j} \mathbf{m}_2]
		+ \mathbb{E}[\mathbf{v}_i^\top \mathbf{m}_2 \mathbf{v}_j^\top \mathbf{m}_2].
	\end{multline*}

	Since every element of $\mathbf{M}_1$ and $\mathbf{M}_2$ follows the standard normal distribution, the following are satisfied.
	First, $\mathbb{E}[m_{1i}m_{ij}] = 1$ if $i = j$ and zero otherwise.
	Second, the second and third elements of the right hand side are zero.
	Third, $\mathbb{E}[\mathbf{v}_i^\top \mathbf{m}_2 \mathbf{v}_j^\top \mathbf{m}_2] = \mathbf{v}_i^\top \mathbf{v}_j$.
	As a result, we have the following equality:
	\begin{equation*}
		\mathbb{E}[(z_i - \mathbb{E}[z_i])(z_j - \mathbb{E}[z_j])] = \beta \mathbb{I}[i = j] + \mathbf{v}_i^\top \mathbf{v}_j,
	\end{equation*}
	where $\mathbb{I}$ is an indicator function that returns one if the condition holds, and zero otherwise.
	This equation is the same as the definition of $\Sigma = \beta \mathbf{I} + \mathbf{V} \mathbf{V}^\top$ in the matrix form.
\end{proof}

\subsection{Proof of Lemma \ref{lemma:equivalence}}
\label{appendix:proof-equivalence}

\begin{proof}
	$\mathrm{tr}(\mathbf{K} \Sigma) = \beta \mathrm{tr}(\mathbf{K}) + \mathrm{tr}(\mathbf{K}\mathbf{E}\mathbf{E}^\top)$ due to the definition of $\Sigma$.
	The cyclic property of a trace makes $\mathrm{tr}(\mathbf{K}\mathbf{E}\mathbf{E}^\top) = \mathrm{tr}(\mathbf{E}^\top\mathbf{K}\mathbf{E})$.
	Thus, $\mathrm{tr}(\mathbf{K} \Sigma) = \beta \mathrm{tr}(\mathbf{K}) + \mathrm{tr}(\mathbf{E}^\top\mathbf{K}\mathbf{E})$, and $\mathrm{tr}(\mathbf{K})$ is a constant.
\end{proof}

\subsection{Proof of Lemma \ref{lemma:time-complexity}}
\label{appendix:proof-time-complexity}

\begin{proof}
	\method consists of an encoder $f$ and two decoders $g_x$ and $g_y$.
	The complexity of $f$ is $O(d^2|\mathcal{V}| + d|\mathcal{E}|)$ assuming the identity feature matrix.
	The complexities of $g_x$ and $g_y$ are $O(md|\mathcal{V}|)$ and $O(cd|\mathcal{V}|)$, respectively.
%	The complexity of the GMRF regularizer is $O(d^2|\mathcal{V}| + d^2|\mathcal{E}| + d^3)$, because $\mathrm{tr}(\mathbf{E}^\top \mathbf{K} \mathbf{E})$ is $O(d^2|\mathcal{V}| + d^2|\mathcal{E}|)$ and $\log|\mathbf{I} + \beta^{-1}\mathbf{E}^\top \mathbf{E}|$ is $O(d^2|\mathcal{V}| + d^3)$.
\end{proof}

\subsection{Proof of Lemma \ref{lemma:space-complexity}}
\label{appendix:proof-space-complexity}

\begin{proof}
	\method consists of an encoder $f$ and two decoders $g_x$ and $g_y$.
	The complexity of $f$ is $O(d|\mathcal{V}| + |\mathcal{E}| + d^2)$ assuming the identity feature matrix.
	The complexities of $g_x$ and $g_y$ are $O(m|\mathcal{V}| + d|\mathcal{V}| + md)$ and $O(c|\mathcal{V}| + d|\mathcal{V}| + cd)$, respectively.
%	The complexity of the GMRF regularizer is $O(d|\mathcal{V}| + d|\mathcal{E}| + d^2)$, as $\mathrm{tr}(\mathbf{E}^\top \mathbf{K} \mathbf{E})$ is $O(d|\mathcal{V}| + d|\mathcal{E}|)$ and the $\log \det$ is $O(d|\mathcal{V}| + d^2)$.
\end{proof}

\section{Details of Stochastic Inference}
\label{appendix:algorithm}

\begin{algorithm}[t]
\caption{Basic version of structured variational inference.}
\begin{algorithmic}[1]
	\Require Adjacency matrix $\mathbf{A}$, diagonal adjacency $\mathbf{D}$, feature $\mathbf{X}$, (optional) one-hot label $\mathbf{Y}$, hyperparameter $\beta$, networks $f_\mu$, $f_\sigma$, $g_x$, and $g_y$, and their parameters $\phi_\mu$, $\phi_\sigma$, $\theta$, and $\rho$, resp.
	\Ensure Updated parameters $\phi_\mu'$, $\phi_\sigma'$, $\theta'$, and $\rho'$
	\State $\mathbf{U}, \mathbf{V} \leftarrow f_\mu(\mathbf{A}; \phi_\mu), f_\sigma(\mathbf{A}; \phi_\sigma)$ \Comment Run encoder functions
	\State $\Sigma \leftarrow \beta \mathbf{I} + \mathbf{V} \mathbf{V}^\top$ \Comment Make the covariance matrix
	\State $\mathbf{M}_1, \mathbf{M}_2 \leftarrow \mathrm{StandardNormal}()$ \Comment Sample random matrices
	\State $\mathbf{Z} \leftarrow \mathbf{U} + \sqrt{\beta} \mathbf{M}_1 + \mathbf{V} \mathbf{M}_2$ \Comment Make latent variables
	\State $\hat{\mathbf{X}}, \hat{\mathbf{Y}} \leftarrow g_x (\mathbf{Z}, \mathbf{A}; \theta), g_y (\mathbf{Z}, \mathbf{A}; \rho)$ \Comment Make predictions
	\State $l_{xy} \leftarrow \sum_i l_x(\hat{\mathbf{x}}_i, \mathbf{x}_i) + \sum_j l_y(\hat{\mathbf{y}}_j, \mathbf{y}_j)$ \Comment Equation \eqref{eq:loss-x} to \eqref{eq:loss-x3}
	\State $\mathbf{K} \leftarrow \mathbf{I} - \mathbf{D}^{-1/2} \adj \mathbf{D}^{-1/2}$ \Comment Equation \eqref{eq:info-matrix}
	\State $l_\mathrm{KLD} \leftarrow 0.5 ( \mathrm{tr}(\mathbf{U}^\top \mathbf{K} \mathbf{U}) + d(\mathrm{tr}(\mathbf{K}\Sigma) - \log|\Sigma|))$  \Comment Equation \eqref{eq:kl-divergence}
	\State$\phi'_\mu, \phi'_\sigma, \theta', \rho' \leftarrow$ Update $\phi_\mu, \phi_\sigma, \theta, \rho$ to minimize $l_{xy} + l_\mathrm{KLD}$
\end{algorithmic}
\label{alg:stochastic}
\end{algorithm}

We present a detailed algorithm of structured variational inference in Algorithm \ref{alg:stochastic}, which performs stochastic sampling described in Section \ref{ssec:method-variational-inference}.
It uses two encoder functions for generating embedding matrices $\mathbf{U}$ and $\mathbf{V}$, respectively, in line 1.
Then, it samples $\mathbf{Z}$ from the Gaussian distribution with the reparametrization trick in lines 2 to 4.
The prediction is done as in the deterministic inference, but the regularizer term works differently in line 8, taking $\mathbf{U}$ and $\Sigma$ as its inputs.
The parameters of all four networks are updated.

\section{Evaluation Metrics}
\label{appendix:evaluation}

We use four metrics for the evaluation of estimated features: recall at $k$ and nDCG at $k$ for binary features, and RMSE and CORR for continuous features.
Categorical features are not included in our datasets in Table \ref{table:datasets}, but we can use classification accuracy as done for evaluating labels.
The symbols used in this section are defined as follows: $n$ is the number of nodes, $d$ is the number of features, $\mathbf{x}_i$ is the true feature vector of node $i$, $\hat{\mathbf{x}}_i$ is the prediction for $\mathbf{x}_i$, $x_{ij}$ is the $j$-th element of $\mathbf{x}_i$, and $\mathbb{I}(\cdot)$ is a binary function that returns one if the condition holds and zero otherwise.

\textbf{Evaluation of binary features.}
We consider the feature estimation for binary features as a ranking problem, which is to find the nonzero elements at each feature vector $\mathbf{x}_i$ of node $i$.
Let $\hat{r}_{ij}$ be the index having the $j$-th largest score in $\hat{\mathbf{x}}_i$.
Then, we use the top $k$ predictions with the largest scores, i.e., $\{ \hat{x}_{il} \mid l = \hat{r}_{i1}, \cdots, \hat{r}_{ik} \}$, at each node $i$, where $k$ is chosen in $\{3, 5, 10, 20, 50\}$.
This is done also in previous work for binary feature estimation \cite{Chen2020} to focus more on the predictive performance of the top predictions.

Recall at $k$ measures the ratio of true entries contained in the top $k$ predictions for each node:
\begin{equation}
	\mathrm{REC}_k(\hat{\mathbf{X}}, \mathbf{X}) =
		\frac{1}{n} \sum_{i=1}^n \sum_{j=1}^k \frac{\mathbb{I}[x_{i, \hat{r}_{ij}} = 1]}{\|\mathbf{x}_i\|_0},
\end{equation}
where $\|\mathbf{x}_i\|_0$ is the number of nonzero entries in $\mathbf{x}_i$.
For instance, recall @ 3 is computed as $2/3$ in the following example:
\begin{align*}
	&\mathbf{x}_i = (0, 0, 1, 1, 1) \\
	&\hat{\mathbf{x}}_i = (0.1, 0.7, 0.2, 0.8, 0.9),
\end{align*}
since $\hat{r}_{i1} = 5$, $\hat{r}_{i2} = 4$, and $\hat{r}_{i3} = 2$, and two of the nonzero entries of $\mathbf{x}_i$ are included in the top 3 predictions with the largest scores.

nDCG at $k$ measures also the quality of order in the top $k$ predictions with respect to information retrieval.
nDCG is computed by normalizing DCG at $k$, which is defined as
\begin{equation}
       \mathrm{DCG}_k(\hat{\mathbf{X}}, \mathbf{X}) = \frac{1}{n} \sum_{i=1}^n \sum_{j=1}^k \frac{\mathbb{I}(x_{i, \hat{r}_{ij}} = 1)}{\log_2(j + 1)}.
\end{equation}

\textbf{Evaluation of continuous features.}
We evaluate predictions for continuous features with simple metrics of RMSE and CORR.
RMSE measures an error between predictions and true features:
\begin{equation}
	\mathrm{RMSE}(\hat{\mathbf{X}}, \mathbf{X}) =
		\frac{1}{n} \sum_{i=1}^n \sqrt{\frac{1}{d} \sum_{j=1}^d (\hat{x}_{ij} - x_{ij})^2}.
\end{equation}

CORR measures how much predictions and true features are correlated, and is defined as follows:
\begin{equation}
	\mathrm{CORR}(\hat{\mathbf{X}}, \mathbf{X}) =
		\frac{1}{d} \sum_{j=1}^d \left( 1 - \frac{\sum_{i=1}^n (\hat{x}_{ij} - x_{ij})^2}{\sum_{i=1}^n (x_{ij} - \bar{x}_j)^2} \right),
\end{equation}
where $\bar{x}_j = \sum_{i=1}^n x_{ij} / n$ is the mean of the $j$-th feature of all nodes.
CORR is higher the better, while RMSE is lower the better.

%\begin{table}
%\caption{%
%	The performance of our \method with different values of $\lambda$ and $\beta$.
%	We report the recall @ $k$ scores with $k=10$.
%}
%\begin{tabular}{cc|ccccc}
%	\toprule
%	$\lambda$ & $\beta$ & Cora & Cite. & Comp. & Photo & Steam \\
%	\midrule
%	0.1 & 0.1 & .1635 & \textbf{.0943} & .0431 & .0438 & .2524 \\
%	0.1 & 1.0 & .1622 & .0932 & .0430 & .0437 & .2528 \\
%	1.0 & 0.1 & \textbf{.1718} & .0925 & \textbf{.0437} & \textbf{.0446} & \textbf{.2565} \\
%	1.0 & 1.0 & .1715 & .0917 & .0436 & \textbf{.0446} & .2557 \\
%	\bottomrule
%\end{tabular}
%\label{table:sensitivity}
%\end{table}

\section{Hyperparameter Settings}
\label{appendix:parameters}

We search the hyperparameters of our \method as follows: the size $d$ of latent variables in $\{256, 512\}$, the dropout probability in $\{0.0, 0.5\}$, the regularization parameters $\lambda$ and $\beta$ in $\{0.01, 0.1, 1.0\}$, and the unit normalization of latent variables in $\{\mathrm{true}, \mathrm{false}\}$.
We also use the Adam \cite{Kingma2015} optimizer with the learning rate $r=0.005$ in Steam and $r=0.001$ in all other datasets.
The early stopping is used with the validation performance, and all of our experiments were done at a workstation with RTX 2080 based on PyTorch.
More detailed information can be found in our official implementation.\footnote{\url{https://github.com/snudatalab/SVGA}}

%Table \ref{table:sensitivity} shows the parameter sensitivity of our \method with respect to $\lambda$ and $\beta$.
%We report only the recall @ $k$ scores with $k=10$ since other evaluation metrics present similar trends.
%\method works well in all settings of $\lambda$ and $\beta$, compared with the baseline approaches in Table \ref{table:feature-estimation}, demonstrating its robustness for the choice of hyperparameters.
%The setting with $\lambda=1.0$ and $\beta=0.1$ performs the best in four of the five datasets.
%We report the test performance in Table \ref{table:sensitivity}, but the trend is the same in the validation data, allowing us to find the best hyperparameters during the training.

\end{document}